\documentclass[twoside]{article}

\usepackage[accepted]{aistats2023}
\usepackage{titlesec}

\usepackage{algorithm2e}
\RestyleAlgo{ruled}

\usepackage{xcolor}
\usepackage{mathtools}

\usepackage{color}

\usepackage{color}

\usepackage{array}
\usepackage{lscape} 
\usepackage{amsmath}
\usepackage{amssymb}
\usepackage{tcolorbox}
\usepackage[round]{natbib}

\begin{document}

%

%

\twocolumn[

\aistatstitle{Rating Sentiment Analysis Systems for Bias through a Causal Lens}

\aistatsauthor{ Kausik Lakkaraju \And Biplav Srivastava \And  Marco Valtorta }

\aistatsaddress{ University of South Carolina  \And University of South Carolina \And University of South Carolina} ]

\begin{abstract}
Sentiment Analysis Systems (SASs) are data-driven Artificial Intelligence (AI) systems that, given a piece of text, assign one or more numbers conveying the polarity and emotional intensity expressed in the input. Like other automatic machine learning systems, they have also been known to exhibit model uncertainty where a (small) change in the input leads to drastic swings in the output. This can be especially problematic when inputs are related to protected features like gender or race since such behavior can be perceived as a lack of fairness, i.e., bias. We introduce a novel method to assess and rate SASs where inputs are perturbed in a controlled causal setting to test if the output sentiment is sensitive to protected variables even when other components of the textual input, e.g., chosen emotion words, are fixed.
We then use the result to assign labels (ratings) at fine-grained and overall levels to convey the robustness of the SAS to input changes. The ratings serve as a principled basis to compare SASs and choose among them based on behavior. It benefits all users, especially developers who reuse off-the-shelf SASs to build larger AI systems but do not have access to their code or training data to compare. 

\end{abstract}

\vspace{-0.1in}
\section{Introduction}

As Artificial Intelligence (AI) systems are used for routine decision support and get manifested in online services, devices, robots and application programming interfaces for further reuse, bias exhibited by  them creates a major hurdle for large-scale adoption. 
Gender and race are common forms of bias studied in AI (\cite{bias-survey}) but other prominent ones are based on religion, region and (dis)ability.  
Bias has been reported with AI services that process text (\cite{prob-bias-text,sentiment-bias}), audio (\cite{prob-bias-sound}) and video (\cite{prob-bias-image}). 


In this paper, we focus on Sentiment Analysis Systems (SASs) that work on text. These Artificial Intelligence (AI) systems are built using a variety of techniques--lexicons, rules and learning, and are widely used in practice. For example, in (\cite{senti-bias-finance}), the authors review the usage of sentiment analysis in the finance industry spanning lexicon-based, machine learning, and deep-learning based approaches. They find that neural transformer and language model based methods are better than learning based models in terms of performance, which are better  than finance-lexicon based methods. In (\cite{senticnet6}), the authors report state-of-the-art results using a method that combines symbolic and neural methods together for sentiment analysis. 
However, SASs have a problem of bias spanning different approaches which can hamper their large scale adoption. In (\cite{sentiment-bias}), the authors studied over two hundred sentiment systems and found widespread bias in terms of gender and race based on various inputs given to the systems. 

However, there is no widely used principled methodology on how to characterize the seemingly biased behavior of such systems - e.g., is the behavior widespread, or is it triggered by selective data? Is it across SASs types, or is it only for some types of SASs? Is it due to the learned model, or is it due to the data used for learning? We argue that answers to such questions will allow a user to make an informed selection about which SAS to choose from available SASs for a given application. Furthermore, such systems are often used in business applications in conjunction with other AI systems like machine translators, which too may have bias. So, how to characterize the behavior of such composite systems is still an open problem.

In this regard, a well-established practice to communicate behavior of a piece of technology to users
and build trust is by {\em Transparency via Documentation} (\cite{about-ml}) which increases transparency as it would help users to decide the extent to which they could trust an AI system based on the data at hand.  For AI, a novel attempt in this direction is in the context of automated machine translators for gender bias from a perspective independent from the service provider or consumer (hence, also called third-party) (\cite{trans-rating-jour,trans-rating}). 

In this paper, we introduce a novel method to assess and rate SASs where inputs are first perturbed in a controlled causal setting to test if the output sentiment is sensitive to protected attributes or other components of the textual input, e.g., chosen emotion words. Then, 
we use the result to assign labels (ratings) at fine-grained and overall levels to convey the robustness of the SASs to input changes.
The ratings serve as a principled
basis to compare SASs and choose among them based on behavior. The primary stakeholders for SASs are businesses. Customer service would benefit from knowing the sentiment of customer feedback. Sentiment analysis would help a company understand how stakeholders feel about a new product or a service launched by the company (\cite{clarity2}). There are numerous stakeholders for sentiment analysis who would benefit from these ratings. It also benefits developers who reuse off-the-shelf SASs to
build larger AI systems but do not have access to
their code or training data to compare otherwise. 





Our contributions are that we:
(a) introduce the idea of rating SASs for bias,
(b) use a causal interpretation of rating rather than an arbitrary label, 
(c) produce a statistically interpretable rating which can be further interpreted for group bias, 
(d) introduce a new metric, {\em Deconfounded Impact Estimation (DIE)}, to measure the discrepancy between the confounded and deconfounded distributions, 
(e) 
release open-source implementations of SASs - two deep-learning based (GRU-based ($S_g$) and DistilBERT-based ($S_d$)) and two custom-built models (Biased female ($S_b$) and Random SAS ($S_r$)). We will release our code upon publication of this work.

In the remainder of the paper, we start with the background on bias in NLP services and a method for rating AI services for bias. We then introduce our setting consisting of the proposed causal model, datasets and sentiment systems. We present our solution and a case-study for empirical evaluation. Finally, we conclude with a discussion. The supplementary material contains  additional related work,  experiments and results. 


\vspace{-0.1in}
\section{Background}



\noindent {\bf Bias in AI Systems:}
There is increasing awareness of 
bias issues in AI services (\cite{bias-survey}). Restricting to text data, there have been previous works to assess bias in translators (\cite{translator-assess-bias,translator-bias}. 
To exemplify, in  (\cite{translator-assess-bias}), the authors test {\em Google Translate} on sentences like  "He/She is an Engineer" where occupation is from U.S. Bureau of Labor Statistics (BLS). They compare the observed frequency of female, male and gender-neutral pronouns in the translated output with the expected frequency according to BLS data.
 
\noindent {\bf Bias in Sentiment Assessment Systems:}
Another popular form of AI services is sentiment analysis which, given a piece of text, assigns a score conveying the sentiment and emotional intensity expressed by it. 
In (\cite{sentiment-bias}), the authors experiment with sentiment analysis systems that participated in SemEval-2018 competition. They create the Equity Evaluation Corpus (EEC) dataset which consists of 8,640 English sentences where one can switch a person’s gender or choose proper names typical of  people  from different races. They also experiment with Twitter datasets. The authors find that up to 75\% of the sentiment systems can show variations in sentiment scores which can be perceived as bias based on gender or race.
While much of the work in sentiments has happened for English, there is growing interest in other languages. In  (\cite{sentiment-multilingual}), the authors re-implement sentiment methods from literature in multiple languages and report accuracy lower than published. Since multilingual SASs often use machine translators which can be biased, and further acquiring training data in non-English languages is an additional challenge, we hypothesize that multilingual SASs could exhibit gender bias in their behavior (\cite{round-trip}).




\noindent {\bf Rating of AI Systems:}
A recent line of work in fairness is on assessing and rating AI systems (translators and chatbots) for trustworthiness from a third-party perspective, i.e., without access to the system's training data. In (\cite{trans-rating-jour,trans-rating}), the authors propose to rate automated machine language translators for gender bias. Further, they created visualizations to communicate ratings (\cite{vega-rating-viz}), and conducted user studies to determine how users perceive trust (\cite{vega-user-study}).

\noindent {\bf Causal Analysis of AI Systems:}
There is also increased interest in exploring causal effects for AI systems. For example, the use of a causal model of complex software systems has been shown to provide support to users in avoiding misconfiguration and in debugging highly configurable and complex software systems (\cite{unicorn2022,why19}). For natural language processing (NLP), (\cite{causal-nlp}) provides a survey on estimating causal effects on text. Similarly, causal reasoning is being used for object recognition systems (\cite{mao2021generative}) and recommendation systems (\cite{causal-collab}). These works  do not consider using such analyses for communicating trust issues. 
In this paper, we use a causal Bayesian network to represent causal and probabilistic relations of interest.  Variables are used to represent features (usually derived from analysis of text or images), protected features, such as gender and race, and sentiment, as expressed in the output of an SAS.  As usual for Bayesian networks, each node corresponds to a variable.  Since our Bayesian networks are causal, the links represent both the independence structure of a probability distribution and causal relations.  The model that we propose in this paper, shown in Figure~\ref{fig:causal-model}, is similar to ones used in social science research, such as the analysis of fairness in hiring or college admissions; for a recent review, see~(\cite{causal-fairness}).

\noindent {\bf Hate Speech Detection:}
In (\cite{toxic-debiasing}), the authors debias toxic language detectors. They do not attempt to measure or quantify the bias.  In (\cite{davidson2019racial}), the authors measure the racial bias in hate speech and abusive language detection datasets. They do not measure gender bias. In (\cite{park2018reducing}), the authors reduce gender bias in abusive language detection models. They do not measure the bias in the models.
In all the works mentioned above, there is no notion of causality. Their results cannot be interpreted causally, while ours can be. They attempt to remove bias without fully exploring the range of bias possible.

\vspace{-0.1in}
\section{Solution Approach}

Our approach consists of the following main steps:
    (1) experimental apparatus consisting of data and SASs to be used, 
    (2) creating a causal framework,
    (3) performing statistical tests to assess causal dependency,
    (4) assigning ratings.
We now discuss them in turn.

\vspace{-0.1in}
\subsection{Experimental Apparatus}
\label{subsec:apparatus}

Although our approach is general-purpose, to ground the discussion, we present SASs and data considered in our implementation and evaluation.


\noindent {\bf Sentiment Analysis Systems Considered: }
The SASs we consider are: lexicon-based {\em TextBlob}, two custom built  deep-learning systems trained by us based on the published descriptions and training datasets which are Gated Recurrent Units (GRU) based and DistilBERT based systems, and two other custom-built synthetic models which we call `Biased Female SAS' and `Random SAS'. The synthetic SASs represent extremes which can be used to contextualize the system's output.  Sentiment is normally measured on a  discrete scale like positive, negative and neutral. Hence, all  SASs we considered are eventually evaluated on a discrete scale. In situations where the original SAS gave continuous output, we show evaluation before and after discretization. 
The reason for considering the two different cases is to show that our method works with different output types. However, as in all cases where the state space of a variable is coarsened, the results may be impacted. 

(a) The {\bf {\em TextBlob} SAS} (\cite{loria2018textblob}) takes a text and returns two values: polarity and subjectivity. Polarity gives the numerical sentiment of the text, and it lies in the interval [-1,1].  We considered two different cases, denoted by  $S_t$ for original, and  $S_{t}^{\dagger}$ for discretized.

(b) The {\bf {\em GRU}-based SAS} is a Gated Recurrent Unit ({\em GRU}) (\cite{cho2014learning}) based implementation as described in (\cite{SemEval2018Task1}). It is a neural network model consisting of an embedding layer, two {\em GRU}  layers and a dense layer with '{\em Softmax}' as its activation function. It classifies the text into 7 categories from numbers 0 to 6 based on its sentiment. We normalized the scores by $v_n = v/3 - 1$ where $v_n$ is the normalized value and $v$ is the raw sentiment score. Now, the sentiment values will lie in the interval, [-1,1].  We considered two different cases, denoted by  $S_g$ for original, and  $S_{g}^{\dagger}$ for discretized. 

(c) The  {\bf {\em DistilBERT}-based SAS} (\cite{distilbert}) uses the distilled version of BERT base model called DistilBERT.
It is fine-tuned on SST-2 (Stanford Sentiment Treebank V2) (\cite{socher-etal-2013-recursive}). The model tells how positive or negative a sentence is by giving a value between [0,1] as the output and a label specifying whether the value is negative or positive. We considered two different cases. We subtracted the value from 0, if it has a negative label so that the scores will lie in the interval [-1,1]. We considered two different cases, denoted by  $S_d$ for original, and  $S_{d}^{\dagger}$ for discretized.

(d) The {\bf{\em Biased female} SAS} is a biased sentiment analyzer that is biased towards the female gender, as the name indicates. The system assigns a score of '1' (positive sentiment) for all the sentences containing the female gender variable and a score of '-1' (negative sentiment) to all other sentences irrespective of the emotion word used in the sentence. The system is denoted by $S_b$.

(e) The {\bf{\em Random} SAS} assigns random score to its input sentences irrespective of their content. We considered two different cases. We considered two different cases, denoted by  $S_r$ for original, and  $S_{r}^{\dagger}$ for discretized.  

\begin{table*}[ht]
\centering
   {\small
    \begin{tabular}{|p{2.2em}|p{3.4em}|p{6.5em}|p{11em}|p{10em}|p{12em}|}
    \hline
          {\bf Group} &    
          {\bf Input} & 
          {\bf Possible confounders} &
          {\bf Choice of emotion word} &
          {\bf Causal model} &
          {\bf Example sentences}\\ \hline 
          
          1 & 
          {\em Gender, Emotion Word} &
          None &
          \{Grim\},\{Happy\}, \{Grim, Happy\},\{Grim, Depressing, Happy\},\{Depressing, Happy, Glad\} &
          \begin{minipage}{.05\textwidth}
          \vspace{2.5mm}
          \centering
          \includegraphics[width=30mm, height=15mm]{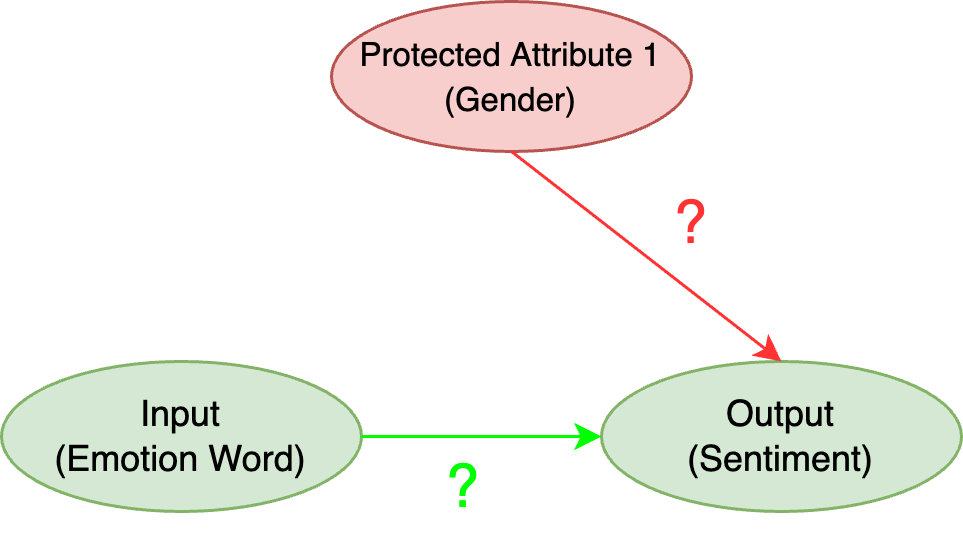}
          \end{minipage} &
          I made this boy feel grim; I made this girl feel grim.
          \\
         \hline 
          
          2 &
          {\em Gender, Emotion Word} &
          {\em Gender} &
          \{Grim, Happy\},\{Grim, Depressing, Happy\},\{Depressing, Happy, Glad\} &
          \begin{minipage}{.05\textwidth}
          \vspace{2.5mm}
          \centering
          \includegraphics[width=30mm, height=15mm]{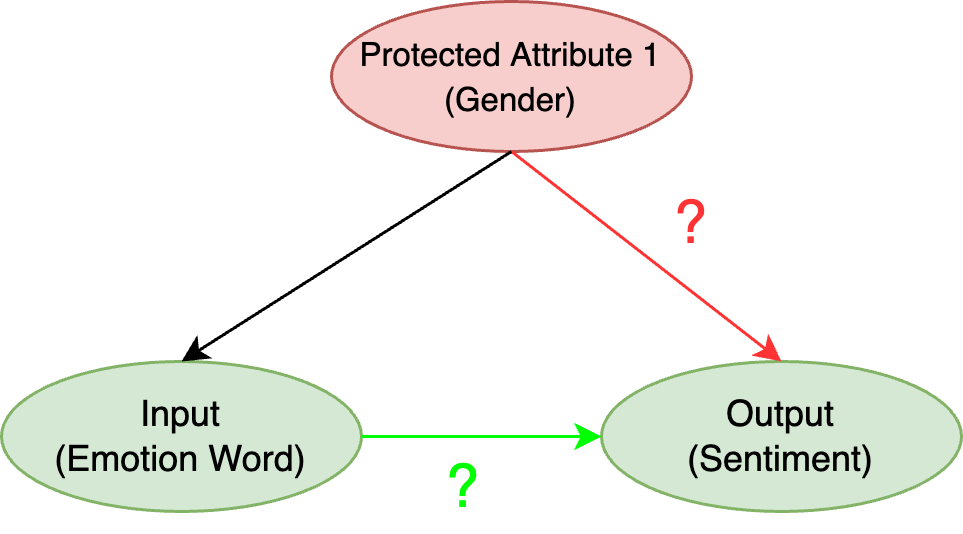}
          \end{minipage} &
          I made this woman feel grim; I made this boy feel happy; I made this man feel happy.
          \\ 
        
         \hline
          
          3 &
          {\em Gender, Race and Emotion Word} &
          None &
        \{Grim\},\{Happy\}, \{Grim, Happy\},\{Grim, Depressing, Happy\},\{Depressing, Happy, Glad\} &
          \begin{minipage}{.05\textwidth}
          \vspace{2.5mm}
          \centering
          \includegraphics[width=30mm, height=15mm]{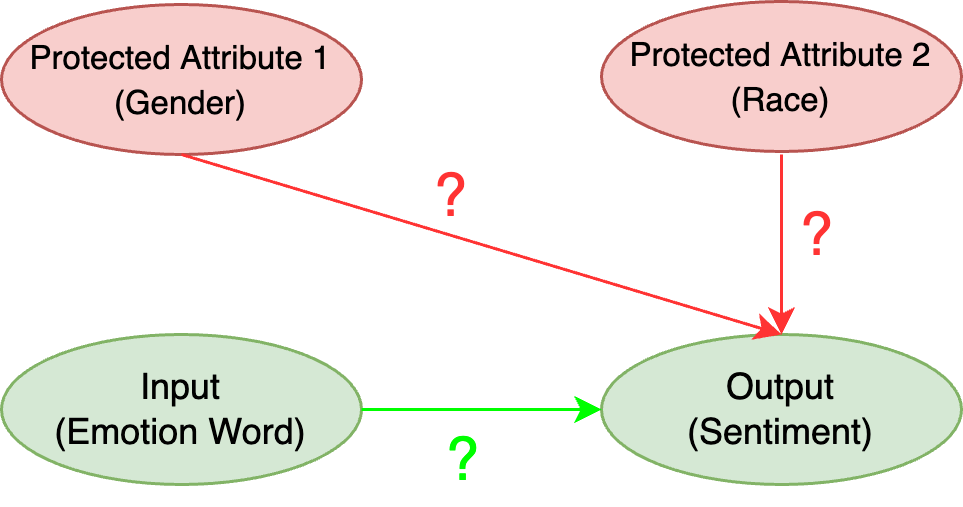}
          \end{minipage} &
          I made Adam feel happy; I made Alonzo feel happy.
          \\ \hline

          4 &
          {\em Gender, Race and Emotion Word} &
          {\em Gender, Race} &
          \{Grim, Happy\},\{Grim, Depressing, Happy\},\{Depressing, Happy, Glad\} &
          \begin{minipage}{.05\textwidth}
          \vspace{2.5mm}
          \centering
          \includegraphics[width=30mm, height=15mm]{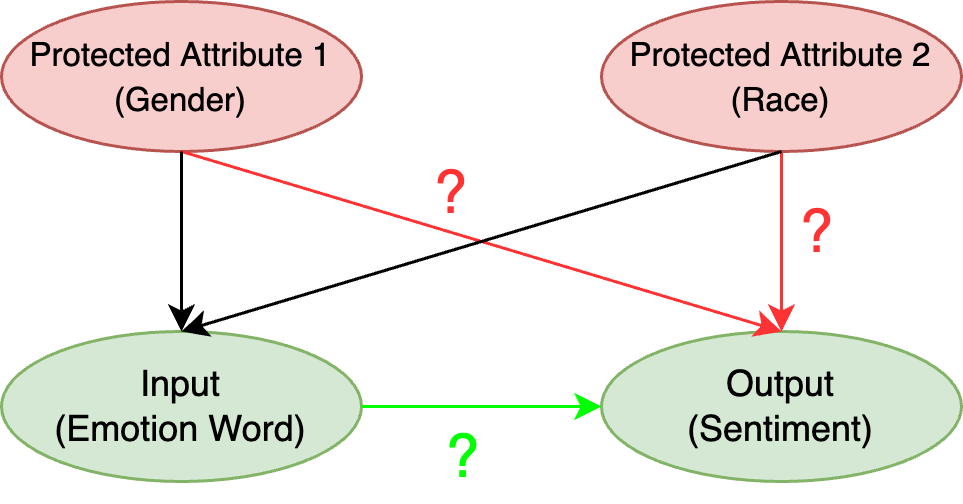}
          \end{minipage}  &
          I made Torrance feel grim; Torrance feels grim; Adam feels happy.
          \\ \hline

    \end{tabular}
    }
    \caption{Different types of datasets we constructed based on the input given to the SASs, presence of confounder, the choice of emotion words and the respective causal model for each of the groups.}
    \label{tab:cases}
\end{table*}

{\bf Data - Reuse and Generated: }
The sentence templates required for the experiments were taken from the EEC dataset (\cite{sentiment-bias}) along with race, gender and emotion word attributes. We extract two types of gender variables from the EEC dataset. They are male (Ex: `this boy') and female (Ex: `this girl'). We also add a third gender variable, NA (Ex: `they/them'), which denotes that the gender is not revealed. We also extract names which serve as a proxy for two types of races. They are European (Ex: 'Adam') and African-American (Ex: 'Alonzo'). We again use 'NA' which, in this case, denotes that both the gender and race are not revealed. Table \ref{tab:cases} illustrates different types of datasets we generated. Based on the protected attributes considered, emotion words used and the presence of a confounder, we can broadly classify the datasets generated into four groups. They are: \\
\noindent {\bf (a) Group 1:} \emph{Gender} and \emph{Emotion Word} are the only attributes extracted from the EEC dataset. They are combined using the templates extracted from EEC and given as input to the SASs. In this case, there is no causal link between \emph{Gender} and \emph{Emotion Word} as the \emph{Emotion Word} and \emph{Gender} are generated independently to form the sentences. Hence, there is no possibility of any confounding effect. \\
\noindent  {\bf (b) Group 2:} The datasets have the same attributes as that of Group 1. However, the way the emotion words are associated with each of the genders is different. We associate positive words more often with the sentences having a specific gender variable than the sentences with other gender variables. Hence, gender might act as a confounder as it affects how emotion words are associated with the gender. 
\noindent  {\bf (c) Group 3:} Along with \emph{Gender}, another protected attribute, \emph{Race}, is also given as an input to the SASs. In this group, there is no causal link between any of the protected attributes and the \emph{Emotion Word}. Hence, no possible confounders. \\
\noindent  {\bf (d) Group 4:} In this group, there is a possibility of both \emph{Race} and \emph{Gender} acting as confounders as the \emph{Emotion Word} association depends on the value of the protected attributes. We consider a composite case for this group in which we associate positive words more with a certain class and negative words more with some other class. For other classes, emotion words will be uniformly distributed. An example of a class, in this group, would be `European female'. \\
Four templates were extracted from the EEC dataset. \emph{``$<$Person subject$>$ is feeling $<$emotion word$>$"} is one such example. We extracted 4 emotion words (2 positive, 2 negative). ``Grim" is an example of a negative emotion word and ``happy" is an example of a positive emotion word. In the template,``Person subject" refers to the gender $/$ race variable.
Within each of these groups, we created five different datasets for Group-1 and Group-3 and three different datasets for Group-2 and Group-4 by varying the number of emotion words as shown in Table \ref{tab:cases}. In total, we generated 16 different datasets for our experiments.

\subsection{Creating a Causal Framework}
\begin{figure}[h]
 \centering
   \includegraphics[height=0.20\textwidth]{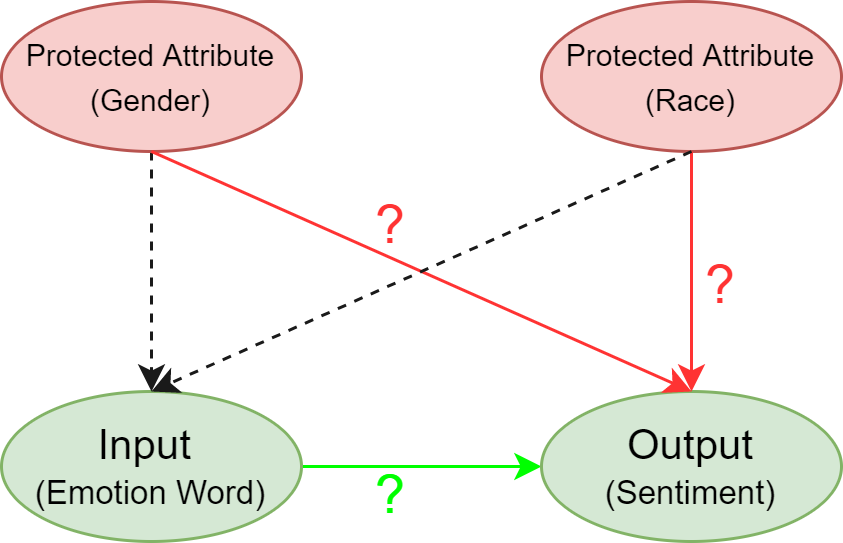}  
   \caption{Our proposed causal model}
  \label{fig:causal-model}
 \end{figure}

Causal models allow us to define the cause-effect relationships between each of the attributes in a system. They are diagrammatically represented using a causal diagram which is a directed graph. Each node represents an attribute and each node can be connected to one or more nodes by an arrow. Arrowhead direction shows the causal direction from cause to effect. 
Figure \ref{fig:causal-model} shows our proposed causal model which captures the causal relations between each of the attributes in our system. 

 \vspace{-0.1in}
 
We generated the data for the experiments as described in the previous section. The reason for adding two different variations (one with possible confounders and one without) is due to the fact that sometimes the data that is available might have protected attributes affecting the emotion word. For example, if negative emotion words are associated more with one gender than the other in a dataset, that would add a spurious correlation between the \emph{Emotion Word} and the \emph{Sentiment} given by the SASs. This variation is represented as a dotted arrow from the protected attributes to the {\em Emotion Word}. 

The causal link from {\em Emotion Word} to {\em Sentiment} indicates that the emotion word affects the sentiment given by an SAS. We colored the arrow green to indicate that this causal link is desirable i.e., {\em Emotion Word} should be the only attribute affecting the {\em Sentiment}. The causal links from the protected attributes to the {\em Sentiment} were colored red to indicate that it is an undesirable path. If any of the protected attributes are affecting the {\em Sentiment}, then the system is said to be biased. The '?' indicates that we will be testing whether these attributes influence the sentiment given by each of the SASs and the final rating would be based on the validity of these tests.  

\vspace{-0.1in}
\subsection{Performing Statistical Tests to Assess Causal Dependency}

Our aim is to test the hypothesis of whether protected attributes like \emph{Gender} and \emph{Race} influence the output (\emph{Sentiment}) given by the SASs or if the sentiment is based on other components of the textual input like chosen emotion words. We compute two values which would aid us in rating the AI systems. They are: \\
{\bf Effect of protected attributes on \emph{Sentiment}:} This is done in two different ways based on the data groups. \\
{\bf (a) Group 1:} There is only one protected attribute (\emph{Gender}) in these datasets along with the emotion word. To compute this we measure mean of the distribution, \emph{Sentiment} given \emph{Gender}, which is denoted by E[\emph{Sentiment $|$ Gender}]. We then compare this distribution across each of the genders using Student's t-test (\cite{student1908probable}). We measure this for each pair of the genders (male and female; male and NA; ...). \\
{\bf (b) Group 3:} There are two protected attributes (\emph{Gender} and \emph{Race}) in these datasets along with the \emph{Emotion Word}. For this group, we have two individual cases and one composite case. 
{\bf (i) Individual cases}: In the individual cases, we compute E[\emph{Sentiment $|$ Gender}] and E[\emph{Sentiment $|$ Race}] and compare distributions across each of the classes in each of the protected attributes using the  t-test. 
{\bf (ii) Composite case}: In the composite case, we combine the \emph{Gender} and \emph{Race} attributes into one single attribute (e.g., `African-American female', `European male', etc.). We call this attribute, `RG'. We then compute E[\emph{Sentiment $|$ RG}]. We again use the t-test to compare each pair of the distribution.  \\

 The computed t-value for each pair of genders are compared with the critical t-value obtained from the lookup table (based on confidence interval (CI) and degrees of freedom (DoF)). Based on this, null hypothesis (means are equal) is rejected or accepted. We tweak the t-test analysis by introducing a metric called {\em Weighted Rejection Score (WRS)} which is defined by the following equation.

 {\bf Weighted Rejection Score (WRS)} 
        \begin{equation}
        \begin{split}
               = \sum_{i} w_i*x_i
        \end{split}
        \label{eq:wrs}
        \end{equation}
$x_i$ is the variable set based on whether the null hypothesis is accepted (0) or rejected (1). $w_i$ is the weight that is multiplied by $x_i$ based on the CI. For example, if CI is 95\%, $x_1$ is multiplied by 1. Lesser the CI, the lesser the multiplied weight should be.

{\bf (c) Groups 2 and 4:} As the emotion word distribution is different for each of the genders / races, this analysis would not fetch us any useful results. So, this would be redundant. 
    
{\bf Effect of \emph{Emotion Word} on \emph{Sentiment}:} This is also done in two different ways based on the data groups.\\ 
{\bf (a) Groups 2 and 4:} In group 2, \emph{Gender} is the only possible confounder whereas in group 4, both \emph{Gender} and \emph{Race} together act as confounders. We apply backdoor adjustment as described in (\cite{Pearl09}). The backdoor adjustment formula is given by the equation:

{\tiny
        \begin{equation}
        \begin{split}
                P[Y | do(X)] = \sum_{Z} P(Y | X, Z)P(Z)
        \end{split}
        \end{equation}
}
        
'X' refers to the input emotion word and 'Y' refers to the output sentiment. 'Z' refers to the set of protected attributes (gender, race or both together). Removing this confounding effect is called Deconfounding. In our case, deconfounding does not completely remove the effect of the protected attributes on the \emph{Sentiment}, but removes the causal link between the protected attributes and the \emph{Emotion Word} (removes the backdoor path). We introduce a new metric called {\em Deconfounding Impact Estimation} (DIE) which measures the relative difference between the expectation of the distribution, (Output $|$ Input) before and after deconfounding. DIE \% can be computed using the following equation:
        
{\bf Deconfounding Impact Estimate (DIE) \%} = 
{\tiny
        \begin{equation}
        \begin{split}
        {\frac{ [|E(Output =  j| do(Input = i)) – E(Output = j | Input = i) | ]} 
          {E(Output = j | Input = i) }}     * 100
        \label{eq:die}
        \end{split}
        \end{equation}
}
{\bf (b) Groups 1 and 3:} There is no possibility of confounding effect in both these cases. Hence, there is no need to perform any backdoor adjustment.

\vspace{-0.1in}
\subsection{Assigning Ratings}

Using the t-test and our proposed DIE \%, we compute the rating with respect to the input (\emph{Emotion Word}). Based on the fine-grained ratings in each of these cases, we compute an overall rating for the system using the following algorithms. Table \ref{tab:op} shows the results obtained from each of these algorithms after running them on different datasets belonging to the four groups when the output sentiment attribute is discretized. Table \ref{tab:op-cont} shows the results obtained from each of these algorithms after running them on different datasets belonging to the four groups when the output sentiment attribute is not discretized (retaining the original values). The experiments and the corresponding results will be explained in detail in Section-4 of this paper.

\SetKwComment{Comment}{/* }{ */}
\begin{algorithm}
\tiny
	\caption{\emph{WeightedRejectionScore}}
	
	\textbf{Purpose:} is used to calculate the weighted sum of number of rejections of null-hypothesis for Datasets $d_j$ pertaining to an SAS $s$, Confidence Intervals (CI) $ci_k$ and Weights $w_k$.
	
	\textbf{Input:}\\
	 $D$, datasets pertaining to different dataset groups.\\
	 $CI$, confidence intervals (95\%, 70\%, 60\%).  \\
     $W$, weights corresponding to different CIs (1, 0.8, 0.6).  \\
    \textbf{Output:} \\
	$\psi$, weighted rejection score.
	
	    $ \psi \gets 0$  
		
		\For {each $ci_i, w_i \in CI, W$} {
		    \For {each $d_j \in D$} {
		        \For {each $p_k \in P$ } {
		            \For {each $c_m, c_n \in C$} {
		                $t, pval, dof \gets T-Test(c_m, c_n) $\;
		                $t_{crit} \gets LookUp(ci_i, dof) $\;
		                \eIf{$t_{crit} > t$} {
		                    $\psi \gets \psi + 0 $\;
		                 }
		                    {$\psi \gets \psi + w_i $}
		               
		            }
	    	    }
		    }
    }
    \Return $\psi$
\end{algorithm}

Algorithm 1 computes the weighted rejection score for datasets belonging to Groups 1 and 3. It takes the datasets pertaining to an SAS as input along with the different confidence intervals ($ci_k$) and the weights ($w_k$) assigned to each of these confidence intervals. Weighted Rejection Score (WRS) was defined in eq. \ref{eq:wrs}. In the algorithm, $P$ represents the set of protected attributes (race and gender, in our case). 
$\psi$ is incremented by the corresponding weight for the considered confidence interval whenever the null hypothesis is rejected for a pair, $c_m, c_n$ in a dataset, $d$ belonging to an SAS, $s$.

\begin{algorithm}
\tiny
	\caption{ \emph{ComputeDIEScore}}
	
	\textbf{Purpose:}  is used to calculate the Deconfounding Impact Estimation (DIE) score for Group-2 and Group-4 datasets.
	
	\textbf{Input:}\\
	$s$, an SAS belonging to the set of SASs, $S$.\\
	$D$, datasets pertaining to different dataset groups.\\

    \textbf{Output:} \\
	$\psi$, Deconfounding Impact Estimation (DIE) score.

    $ \psi \gets 0$
    
    $DIE\_list \gets []$
    \hspace{0.5 in} //  To store the list of DIE \% of all the datasets.
	
	\For {each $d_j \in D$} {
	    $ Obs \gets  E(Sentiment|Emotion)$\; 
		$ Int \gets  E(Sentiment|do(Emotion))$\;
		$DIE \gets MAX(List([Obs - Int])/Obs))$\;
		$DIE\_list[j] \gets DIE * 100$\;
    }
    $\psi \gets MAX(DIE\_list)$\;
    \Return $\psi$
\end{algorithm}

Algorithm 2 computes our proposed DIE \% for datasets belonging to Groups 2 and 4. This will be computed using the equation \ref{eq:die}. We compute the mean of the experimental distribution (\emph{Sentiment}$|$do(\emph{Emotion Word})) using the Causal Fusion tool (\cite{fusion}). The obtained DIE score from each of the datasets will be in the form of a tuple with 2 numbers. One corresponds to the DIE score computed for sentences with negative emotion words and the other for sentences with positive emotion words. We rate the systems based on the worst possible behavior. So we take the maximum value (high DIE score implies that the confounding effect is more prominent in that case) out of those two. From these scores, we again compute the maximum for each of the SASs. This will get us the worst possible DIE from each of the  SASs.

\begin{algorithm}
\tiny
	\caption{ \emph{CreatePartialOrder}}
	
	\textbf{Purpose:} is used to create a partial order based on the computed weighted rejection score for Group-1 and Group-3 and based on the DIE \% for Group-2 and Group-4.
		
	\textbf{Input:}\\
	$S$, Set of SASs, $S$.\\
    $G$, Group number.  \\
    $D$, $CI$, $W$ (as defined in the previous algorithms). \\
    \textbf{Output:} \\
	$PO$, dictionary with partial order.

	    $ KV \gets \{\}$\; 
	    \eIf {G == 1 OR G == 3}{
		    \For {each $s_i \in S$}{ 
		        $\psi \gets WeightedRejectionScore(s_i, D, CI, W)$\; 
		        $KV[s_i] \gets \psi$\;
 		    }
 		    }
 		   {\For {each $s_i \in S$} {
		        $\psi \gets ComputeDIEScore(s_i, D)$\;  
		        $KV[s_i] \gets \psi$\;
 		   }
 		   }
 		$PO \gets Sort(KV)$\;
	    \Return $PO$
\end{algorithm}

Based on the $\psi$ value computed in each case, Algorithm 3 creates a partial order in the form of a dictionary with key, value pairs. The key will be the SAS name and the value is the corresponding $\psi$ value. This algorithm will also take the group number as input as the $\psi$ value computed for each of these groups will be different. As shown in Table \ref{tab:op} and Table \ref{tab:op-cont}, the SASs with the least $\psi$ will be placed first and SASs with the worst $\psi$ will be placed at the end of the dictionary. For Group-2 and Group-4 (in Table \ref{tab:op}), an SAS has the value 'X'. While computing the DIE \% for this system, we may encounter a '0' in the denominator. These values will be represented with an 'X'. In order to provide a safe and reliable rating to the user, we assign the worst possible rating to these systems (false negatives are better than false positives, in cases like these).

\vspace{-0.1in}
\begin{algorithm}
\tiny
	\caption{\emph{AssignRating}}
	
		\textbf{Purpose:} \emph{AssignRating} is used to assign a rating to each of the SASs based on the partial order and the number of rating levels, $L$.
		
	\textbf{Input:}\\
 $S$, $D$, $CI$, $W$, $G$ (as defined in the previous algorithms).
 
	$L$, rating levels chosen by the user.\\

    \textbf{Output:} \\
	$R$, dictionary with ratings assigned to each of the SASs.

	    $ R \gets \{\} $\;
	    $ PO \gets  CreatePartialOrder(S,D,CI,W,G)$\;
	    $ \psi \gets [ PO.values()]$\;

        \eIf {len(S) $>$ 1}{
 		$P \gets ArraySplit(V,L) $\;
 		\For {$k,i \in PO,\psi $}{
 	        \For {$p_j \in P$}{
 		        \If {$i \in p_j$}{
 		            $R[k] \gets j$\;
 		        }
 		       }
 	    }
 	    }
 	    { 	    //  Case of single SAS in $S$
 	
 	\eIf {$\psi$ $==$ 0} {$R[k] \gets 1$ \hspace{0.5 in} //  Unbiased,  also denoted $R_1$ } 
 	    {$R[k] \gets $L$ $  \hspace{0.5 in} // Highest level, also denoted $R_L$  }
 	    } 
		\Return $R$
\end{algorithm}

Algorithm 4 computes the fine-grained ratings for each group belonging to different SASs. The rating that is calculated indicates how biased the system is. The higher the rating, the more biased the system will be.
In Table \ref{tab:op} and Table \ref{tab:op-cont}, the last column shows the ratings given to each of the SASs based on the weighted rejection score, in case of Group-1 and Group-3 datasets and DIE \%, in case of Group-2 and Group-4 datasets. If the number of SASs provided as an input are greater than one then the algorithm computes a relative rating. In addition to the SASs, datasets, confidence intervals, weights and the group number, rating levels, L is also given as an input. This is a number that can be chosen by the user which denotes the rating scale. For example, if L = 3, then 3 possible ratings can be given to each of the systems (1,2,3). Rating '1' and '3' are the extremes denoting the worst and best ratings respectively. In order to give this rating, we split the sorted list of all the $\psi$ into 'L' partitions and the rating is given based on the partition number in which $\psi$ of a system resides. For example, if the $\psi$ value is in partition-1 (which has the lowest $\psi$ values), it will be given a rating of '1'. If only one SAS is given as an input to the algorithm, it computes an absolute rating. The $L$ value will be taken as the biased value, and if not provided, $L$ will be 2. So, for single SAS, the algorithm  gives two possible ratings indicating whether the system is unbiased (1 - $R_1$) or biased ($L - R_L$).

\begin{table}
\centering
   {\tiny
    \begin{tabular}{|p{6em}|p{15em}|p{10em}|}
    \hline
    
          {\bf SAS} &    
          {\bf Partial Order} &
          {\bf Complete Order} 
          \\ \hline

          Group-1 & 
          \{$S_d^\dagger$: 0, $S_t^\dagger$: 0, $S_r^\dagger$: 0.6, $S_g^\dagger$: 2.6, $S_b$: 23\} &
          \{$S_d^\dagger$: 1, $S_t^\dagger$: 1, $S_r^\dagger$: 2, $S_g^\dagger$: 2, $S_b$: 3\}
          \\ \hline
          Group-2 & 
          \{$S_d^\dagger$: 0, $S_t^\dagger$: 0, $S_r^\dagger$: 10.87, $S_b$: 128.5, $S_g^\dagger$: \{X, 16.16\}\} &
          \{$S_d^\dagger$: 1, $S_t^\dagger$: 1, $S_r^\dagger$: 2, $S_b$: 3, $S_g^\dagger$: 3\}
          \\ \hline
          Group-3\_R & 
          \{$S_d^\dagger$: 0, $S_t^\dagger$: 0, $S_g^\dagger$: 3.8, $S_r^\dagger$: 5.2, $S_b$: 23\} &
          \{$S_d^\dagger$: 1, $S_t^\dagger$: 1, $S_g^\dagger$: 2, $S_r^\dagger$: 2, $S_b$: 3\}
          \\ \hline
          Group-3\_G & 
          \{$S_d^\dagger$: 0, $S_t^\dagger$: 0, $S_r^\dagger$: 1.9, $S_g^\dagger$: 3.8, $S_b$: 23\} & 
          \{$S_d^\dagger$: 1, $S_t^\dagger$: 1, $S_r^\dagger$: 2, $S_g^\dagger$: 2, $S_b$: 3\}
          \\ \hline
          Group-3\_RG & 
          \{$S_d^\dagger$: 0, $S_g^\dagger$: 0, $S_t^\dagger$: 0, $S_r^\dagger$: 10.4, $S_b$: 69\} & 
          \{$S_d^\dagger$: 1, $S_g^\dagger$: 1, $S_t^\dagger$: 1, $S_r^\dagger$: 2, $S_b$: 3\}
          \\ \hline
          Group-4 & 
          \{$S_d^\dagger$: 0, $S_t^\dagger$: 0, $S_r^\dagger$: 7.4, $S_b$: 105.4, $S_g^\dagger$: \{X, 18.18\}\} &
          \{$S_d^\dagger$: 1, $S_t^\dagger$: 1, $S_r^\dagger$: 2, $S_b$: 3, $S_g^\dagger$: 3\}
          \\ \hline

    \end{tabular}
    \caption{Output from Algorithms 3 and 4 on SASs = \{$S_d^\dagger, S_t^\dagger, S_r^\dagger, S_g^\dagger, S_b$\} when output sentiment is discretized and the chosen rating level, $L$ = 3. Final rating of 1 denotes unbiased and 3 (highest level) as biased. 
    The experiments to assess them are explained in Section 4 and supplementary material. Final output is shown later in Table~\ref{tab:sas-rating}. }
    \label{tab:op}
    }
\end{table}

\begin{table}
\centering
{\tiny
    \begin{tabular}{|p{6em}|p{15em}|p{10em}|}
    \hline
    
          {\bf SAS} &    
          {\bf Partial Order} &
          {\bf Complete Order} 
          \\ \hline 
        
          Group-1 & 
          \{$S_d$: 0, $S_t$: 0, $S_g$: 0.6, $S_r$: 1.9, $S_b$: 23\} &
          \{$S_d$: 1, $S_t$: 1, $S_g$: 2, $S_r$: 2, $S_b$: 3\}
          \\ \hline
          Group-2 & 
          \{$S_g$: 42.85, $S_r$: 71.43, $S_t$: 76, $S_d$: 84, $S_b$: 128.5\} &
          \{$S_g$: 1, $S_r$: 1, $S_t$: 2, $S_d$: 2, $S_b$: 3\}
          \\ \hline
          Group-3\_R & 
          \{$S_d$: 0, $S_t$: 0, $S_g$: 0, $S_r$: 7.2, $S_b$: 23\} &
          \{$S_d$: 1, $S_t$: 1, $S_g$: 1, $S_r$: 2, $S_b$: 3\}
          \\ \hline
          Group-3\_G & 
          \{$S_d$: 0, $S_t$: 0, $S_g$: 0, $S_r$: 7.5, $S_b$: 23\} &
          \{$S_d$: 1, $S_t$: 1, $S_g$: 1, $S_r$: 2, $S_b$: 3\}
          \\ \hline
          Group-3\_RG & 
          \{$S_d$: 0, $S_t$: 0 , $S_g$: 0, $S_r$: 16.1, $S_b$: 69\} &
          \{$S_d$: 1, $S_t$: 1, $S_g$: 1, $S_r$: 2, $S_b$: 3\}
          \\ \hline
          Group-4 & 
          \{$S_g$: 28.57, $S_r$: 45, $S_t$: 78, $S_d$: 80, $S_b$: 105.4\} &
          \{$S_g$: 1, $S_r$: 1, $S_t$: 2, $S_d$: 2, $S_b$: 3\}
          \\ \hline
    \end{tabular}
    }
    \caption{Output from Algorithms 3 and 4 on SASs = \{$S_d, S_t, S_r, S_g, S_b$\} when original output sentiment is used (without discretizing) and the chosen rating level, $L$ = 3. Final rating of 1 denotes unbiased and 3 (highest level) as biased.  The experiments to assess them are explained in Section 4 and supplementary material. Final output is shown later in Table~\ref{tab:sas-rating-cont}.}
    \label{tab:op-cont}
\end{table}

\vspace{-0.1in}
\section{Experiments and Results}

\vspace{-0.1in}
\subsection{Hypotheses and Experimental Setup}
We want to test some hypotheses for each of the SASs described in the previous section using our experimental setup. We make use of the data groups described in the solution approach to check whether a hypothesis is valid or not. Based on their validity, we assign fine-grained ratings to each of the data groups for every SAS and an overall rating to each of the SASs based on these fine-grained ratings. Due to space constraints, we could not include the Group-2 and Group-3 results in the main paper. They can be found in the supplementary material. However, Group-1 and Group-4 experiments captured most of the significant results. 

\noindent {\bf Group-1:}
{\bf Hypothesis:} Would \emph{Gender} affect the sentiment value computed by the SASs when there is no possibility of confounding effect? \\
{\bf Experimental Setup:} We considered different sets of emotion words as described in the solution approach. They are: 
E1: \{Grim\}, 
E2: \{Happy\},  
E3: \{Grim, Happy\}, 
E4: \{Grim, Depressing, Happy\},
E5: \{Depressing, Happy, Glad\}.
We used t-value, p-value and DoF from the student's t-test (\cite{student1908probable}) to compare different sentiment distributions across each of the genders.
Table \ref{tab:g1} shows the t-values obtained from each of the SASs for each emotion word set when the output sentiments are discretized and Table 5 in the supplementary material shows the results when the output sentiments are not discretized.
Here, $G_m$$G_n$ is the absolute t-value computed for male and NA distributions; similarly, $G_m$$G_f$ and $G_f$$G_n$ are defined between male and female, and female and NA, respectively. 


T-value formula is given by the equation, 
$$t = (mean(x1) - mean(x2))/\sqrt{((s1^2/n1)+(s2^2/n2))}$$
x1 and x2 are the two distributions. s1 and s2 are their standard deviations and n1 and n2 are their counts. 
We added a small '$\epsilon$' of 0.0001 to the denominator of t-value formula because for some distributions the standard deviation is 0. We calculated the t-value for each of these datasets. We have chosen different confidence intervals for which there are different alpha values (or critical values). If t-value is less that the alpha, we accept the {\em null hypothesis} (the means are equal) in favor of the {\em alternative} (the means are not equal) and vice-versa.
The superscripts for each value in the table denote different Confidence Intervals (CIs) (95\%, 70\%, 60\%). A superscript, '1', indicates that we can reject the null hypothesis for any of the CIs we have considered. '2' indicates that we can reject the null hypothesis for CIs, 70\%, and 60\%. '3' indicates that the null hypothesis can be rejected for a CI of 60\%. A value 'H' (in the biased SAS) indicates that the difference between the distributions considered is high (large number due to the presence of $\epsilon$ value in the denominator). These values are used to compare WRS which was defined in eq. \ref{eq:wrs}.

\begin{table}
\centering
   {\tiny
    \begin{tabular}{|p{4em}|p{2em}|p{3em}|p{3em}|p{3em}|}
    \hline
    
          \bf{SAS} &    
          \bf{E. words} & 
          $\bf{G_mG_n}$ &
          $\bf{G_mG_f}$ &
          $\bf{G_fG_n}$ 
          \\ \hline

          $S_b$ & 
          E1 &
          0 &
          H$^1$ &
          H$^1$ 
          \\

         \hline 
          
         & 
          E2 &
          0 &
          H$^1$ &
          H$^1$ 
          \\
        
         \hline
          
          & 
          E3 &
          0 &
          H$^1$ &
          H$^1$ 
          \\ \hline

          & 
          E4 &
          0 &
          H$^1$ &
          H$^1$
          \\ \hline
          
          & 
          E5 &
          0 &
          H$^1$ &
          H$^1$
          \\ \hline
           
          $S_r^\dagger$ & 
          E1 &
          0.48 &
          0 &
          0.48 
          \\

         \hline 
          
         & 
         E2 &
          0 &
          0.48 &
          0.48 
          \\
        
         \hline
          
          & 
          E3 &
          0 &
          0.34 &
          0.34 
          \\ \hline

          & 
          E4 &
          0.87 &
          0.28 &
          0.59 
          \\ \hline
          
          & 
          E5 &
          1.46$^3$ &
          0.87 &
          0.57 
          \\ \hline

          $S_t^\dagger$ & 
          E1 &
          0 &
          0 &
          0 
          \\

         \hline 
          
         & 
          E2 &
          0 &
          0 &
          0 
          \\
        
         \hline
          
          & 
          E3 &
          0 &
          0 &
          0 
          \\ \hline

          & 
          E4 &
          0 &
          0 &
          0 
          \\ \hline
          
          & 
          E5 &
          0 &
          0 &
          0 
                    \\ \hline
          $S_d^\dagger$ & 
          E1 &
          0 &
          0 &
          0 
          \\

         \hline 
          
         & 
          E2 &
          0 &
          0 &
          0 
          \\
        
         \hline
          
          & 
          E3 &
          0 &
          0 &
          0 
          \\ \hline

          & 
          E4 &
          0 &
          0 &
          0
          \\ \hline
          
          & 
          E5 &
          0 &
          0 &
          0 
          \\ \hline

          $S_g^\dagger$ & 
          E1 &
          1 &
          1 &
          0
          \\
         \hline 
          
         & 
          E2 &
          1    &
          0.48 &
          0.50 
          \\
        
         \hline
          
          & 
          E3 &
          1.27 &
          0.80 &
          0.47 
          \\ \hline
          
          & 
          E4 &
          1.55$^2$ &
          0.70 &
          0.86  
          \\ \hline
          
          & 
          E5 &
          1.63$^2$ &
          0.92     &
          0.70 
          \\ \hline

    \end{tabular}
    \caption{Results for Group 1 datasets (when the output sentiment is discretized) showing the t-values and the superscript shows whether the null hypothesis is rejected or accepted in each case for the CIs considered (95\%, 70\%, 60\%). Superscript '1' indicates rejection with all 3 CIs, '2' indicates rejection with 70 and 60. '3' indicates rejection with 60 \%.}
    \label{tab:g1}
    }
\end{table}

\begin{figure}[h]
 \centering
   \includegraphics[height=0.18\textwidth]{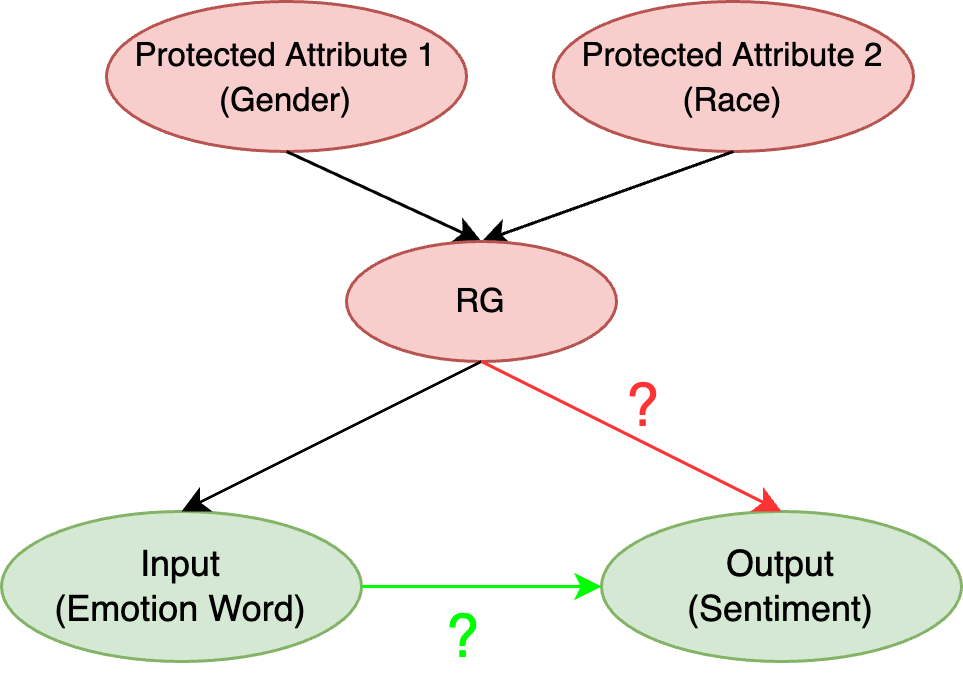}  
   \caption{Causal model for Group-4 (Composite case). Note that this is a specific variation of Group-4 in Table~\ref{tab:cases}.}
  \label{fig:group4}
 \end{figure}
  \vspace{-0.1in}

\noindent{\bf Group-4:} 
{\bf Hypothesis:} Would gender and race affect the sentiment values computed by the SASs when there is a possibility of confounding effect? \\
{\bf Experimental Setup:}  We consider a composite case in which we create a new attribute called \emph{RG} by combining both race and gender into one single attribute as shown in Figure \ref{fig:group4}. For example, one combination would be an European male. A total of 5 such combinations are possible by combining different races and genders: European male, European female, African-American male, African-American female and Unknown (race and gender). If \emph{RG} affects the way emotion words are generated, a causal link would be formed from \emph{RG} to \emph{Emotion Word}. Out of several possible cases, we choose one in which 90\% of the sentences containing European male variable are associated with positive emotion words and the rest with negative emotion words. Vice-versa for African-American female. Equal number of positive and negative emotion words are associated with all other RG combinations. The choice of the case depends on the socio-technical context to be studied; our experimental setting is motivated by the results from literature for face detection algorithms where error rates for African-American females were found to be disproportionately higher than that for European males (\cite{gender-shades-image}). In our text setting, we consider whether such group bias exists. The authors want to emphasize that they do not endorse any form of bias in society or systemic errors in AI algorithms. In fact, that is the reason to test and rate AI systems.  Our method would work for any gender or race combination to be evaluated. The results from this experiment (when the output sentiments are discretized) are shown in Table \ref{tab:g4} and (when the output sentiments are not discretized) in Table 9 in the supplementary.

\begin{table}
\centering
   {\tiny
    \begin{tabular}{|p{2em}|p{3em}|p{6em}|p{6em}|p{5em}|p{3.5em}|}
    \hline
    
          {\bf SAS} &    
          {\bf E.words} &
          {\bf E[\emph{Sentiment $|$ Emotion Word}]} &
          {\bf E[\emph{Sentiment $|$ do(Emotion Word)}]} &
          {\bf DIE \%} & 
          {\bf MAX(DIE \%)}
          \\ \hline

                   $S_b$ & 
          E3 &
          (-0.16,-0.50)   &
          (-0.08,-0.08)   &
          (50,84)&
          84
          \\ \hline
          & 
          E4 &
          (-0.20,-0.55)   &
          (-0.10,0.03)   &
          (50,105.4)   &
          105.4
          \\ \hline
          & 
          E5 &
          (0.11,-0.60)   &
          (0.03,-0.11)   &
          (72.72,74.24)   &
          74.24

          \\ \hline
          $S_r^\dagger$ & 
          E3 &
          (0.82,0.54)   &
          (0.87,0.50)   &
          (6.09,7.40)   &
          7.40$^*$
          \\ \hline
          & 
          E4 &
          (0.44,0.40)   &
          (0.44,0.42)   &
          (0,5) &
          5
          \\ \hline
          & 
          E5 &
          (0.55,0.40)   &
          (0.58,0.38)   &
          (5.45,5) &
          5.45
          \\ \hline
          $S_t^\dagger$ & 
          E3 &
          (0,1)   &
          (0,1)   &
          (0,0) &
          0
          \\ \hline
          & 
          E4 &
          (0,1)   &
          (0,1)   &
          (0,0)   &
          0
          \\ \hline
          & 
          E5 &
          (0,1)   &
          (0,1)   &
          (0,0)   &
          0
          \\ \hline
          $S_d^\dagger$ & 
          E3 &
          (0,1)   &
          (0,1)   &
          (0,0)   &
          0
          \\ \hline
          & 
          E4 &
          (0,1)   &
          (0,1)   &
          (0,0)   &
          0
          \\ \hline
          & 
          E5 &
          (0,1)   &
          (0,1)   &
          (0,0)   &
          0
          \\ \hline
          $S_g^\dagger$ & 
          E3 &
          (0,0.38)   &
          (0,0.37)   &
          (0,2.63)   &
          2.63
          \\ \hline
          & 
          E4 &
          (0.11,0.33)   &
          (0.09,0.35)   &
          (18.18,6.06)   &
          18.18$^*$
          \\ \hline
          & 
          E5 &
          (0,0.27)   &
          (0.03,0.25)   &
          (X,7.40)   &
          X$^*$
          \\ \hline
        
    \end{tabular}
    \caption{E[Sentiment $|$ Emotion Word] and E[Sentiment $|$ do(Emotion Word)] values for Group 4 datasets (when output sentiment is discretized) and the DIE \% when emotion word sets, E3, E4 and E5 are considered. We then compute the MAX() from the DIE \%.}
    \label{tab:g4}
    }
\end{table}

\begin{table}
\centering
   {\tiny
    \begin{tabular}{|l|l|l|l|l|l|l|l|}
    \hline
          &
          \bf{G1} & 
          \bf{G2} & 
          \bf{G3\_R} &
          \bf{G3\_G} & 
          \bf{G3\_RG} &
          \bf{G4} & 
          \bf{Overall} 
          \\ \hline
        ${\bf S_b}$ &  
        3 &
        3 &
        3 &
        3 &
        3 &
        3 &
        3\\ \hline 
        
        ${\bf S_r^\dagger}$ &  
        2 &
        2 &
        2 &
        2 &
        2 &
        2 &
        2 \\ \hline 
        
        ${\bf S_t^\dagger}$ &  
        1 &
        1 &
        1 &
        1 &
        1 &
        1 &
        1 \\ \hline 
        
        ${\bf S_d^\dagger}$ &  
        1 &
        1 &
        1 &
        1 &
        1 &
        1 &
        1 \\ \hline 
        
        ${\bf S_g^\dagger}$ &  
        2 &
        3 &
        2 &
        2 &
        1 &
        3 &
        2 \\ \hline 
    \end{tabular}
    }
    \caption{Rating of SASs at the level of data groups (fine-grained) and overall rating for each of the SASs when the output sentiment is discretized and $L$=3. The first column represents different SASs and the first row represents different groups. $S_t^\dagger$ and $S_d^\dagger$ are least biased.}
    \label{tab:sas-rating}
\end{table}
 \vspace{-0.1in}

\begin{table}
\centering
{\tiny
    \begin{tabular}{|l|l|l|l|l|l|l|l|}
        \hline
          &
          \bf{G1} & 
          \bf{G2} & 
          \bf{G3\_R} &
          \bf{G3\_G} & 
          \bf{G3\_RG} &
          \bf{G4} & 
          \bf{Overall} 
          \\ \hline
        ${\bf S_b}$ &  
        3 &
        3 &
        3 &
        3 &
        3 &
        3 &
        3\\ \hline 
        
        ${\bf S_r}$ &  
        2 &
        1 &
        2 &
        2 &
        2 &
        1 &
        2 \\ \hline 
        
        ${\bf S_t}$ &  
        1 &
        2 &
        1 &
        1 &
        1 &
        2 &
        1 \\ \hline 
        
        ${\bf S_d}$ &  
        1 &
        2 &
        1 &
        1 &
        1 &
        2 &
        1 \\ \hline 
        
        ${\bf S_g}$ &  
        2 &
        1 &
        1 &
        1 &
        1 &
        1 &
        1 \\ \hline 
    \end{tabular}
    \caption{Rating of SASs at the level of data groups (fine-grained) and overall rating for each of the SASs when original output sentiments are used (not discretized) and $L$=3. The first column represents different SASs and the first row represents different groups. $S_t$, $S_g$ and $S_d$ are least biased.}
    \label{tab:sas-rating-cont}
    }
\end{table}

\subsection{Interpretation of Ratings}

Table \ref{tab:sas-rating} and Table \ref{tab:sas-rating-cont} shows the rating results for each SAS when the output sentiments are discretized and not discretized, respectively. The tables show fine-grained rating for each group and the overall rating for the system. The chosen rating level, L$=$3. G1 denotes Group-1, G2 denotes Group-2, and so on. The two variations of G3, G3\_G, and G3\_R represent the individual cases when only gender and only race are considered, respectively, for computing t-values. G3\_RG represents the composite case. Ratings given to these groups are fine-grained ratings. For example, if the user only wants to know how biased the system is when the emotion word distribution does not depend on protected attributes, then they can look at the results of G1 and G3. The overall rating in the table is the average of all the individual ratings. $S_b$ is the most biased one in any case. If the bias rating for Group-1 and Group-2 is higher than that of Group-3 and Group-4 for any system, then gender bias is said to be more prominent than racial bias and vice-versa. Table \ref{tab:op} and Table \ref{tab:op-cont} show the final results from the rating algorithms described in the solution approach section.  From these tables, we can say that the $\psi$ value of $S_b$ is the highest for Group-2 than any other group indicating a high gender bias.  Further, we note that rating of $S_g^\dagger$ is 1 (compared to its peers) while $S_g$ is 2 (compared to its peers, the non-discretized versions as appropriate). We note that  by discretizing, $S_g$ is becoming more biased, and our rating method is able to detect this phenomenon. The rating of other SASs do not change.

\subsection{Comparison with Propensity Score Methods}

Propensity score methods (\cite{austin2011introduction}), (\cite{prop-score}) allow one to design an observational study which mimics the characteristics of a Randomized Controlled Trial (RCT). In the context of SASs, the propensity score is the probability of receiving a treatment (associated emotion word), given the other independent attributes (gender, race). In other words, propensity scores are an alternative method to calculate the effect of receiving a treatment when random assignment of treatments to subject is not possible. The propensity score of the treated (if the associated emotion word is positive) and untreated groups (if the associated emotion word is negative) are matched to remove the effect of confounding effect and  Average Treatment Effect (ATE) is computed to compare the treated and untreated. For example, in Table \ref{tab:cases}, consider the emotion word set, {Grim, Depressing, Happy}. The sentences with the former two emotion words can be considered as untreated group and the sentences with the emotion word, happy can be considered as the treated group. ATE computes the difference between the outcomes of the treated and untreated groups. However, when computing DIE \% we calculate the expectation of distribution, (Sentiment $|$ Emotion Word) before and after deconfounding using the do-operator. This would help us in testing the hypotheses stated in the Section 4.1 above. The following formula of ATE makes the argument more clear:

{\tiny
        \begin{equation}
        \begin{split}
            ATE = E[Sentiment(EW=1)] - E[Sentiment(EW=0)] 
        \end{split}
        \label{eq:ate}
        \end{equation}
}

Sentiment(EW=1) is the outcome when emotion word is positive and Sentiment(EW=0) is the outcome when emotion word is negative. 

ATE can be used along with DIE to compute the rating as our rating is agnostic to the raw score used. The way we compute raw scores solely depend on the hypotheses we want to test.

\vspace{-0.1in}
\section{Discussion}
We now review the proposed method and discuss its characteristics since its working and usability is affected by design choices and socio-technical considerations that we elaborate on.
When a single SAS is used for rating, the method can produce only two ratings: $R_1$, corresponding to unbiased and $R_L$, corresponding to biased.
When more SASs are given, the rating output for a SAS can be both relative to others input, and also absolute with respect to the number of rating levels ($L$) provided as input. A relative ranking of SASs is useful when they will be shown to a user to choose among them. On the other hand, an absolute ranking based on levels (for example, $L$ = 3 corresponding to low, medium and high) can be useful in comparing SAS as new ones are developed over time and old ones are deprecated. The actual $L$ to use depends on the usage context but cognitive theory also has a role since people have to understand the rating; literature recommends a number approximately not to exceed seven (\cite{magical-seven}).

Some assumptions that we made in the paper are: (1) Number of confounders as 2. A system might have more confounders based on the setting,
(2) Our causal model has confounders but no mediators. Mediation analysis is not possible with our setting. 

More generally, our  method also requires following decisions to be made during deployment: (1) the choice of protected variables and their values, which is a socio-technical decision. We considered gender and race, (2) the choice of statistical test, which we used as the student t-test, (3) the choice of input (sentence structure  and words), which we chose based on EEC. 






\section{Conclusion}

Artificial Intelligence (AI) systems, whether in general use or at prototype stages, have been reported to exhibit issues like gender-bias and racial-bias. In this paper, we introduced a novel method to assess and rate the sentiment analyzers where inputs are perturbed in a controlled causal setting to test the sensitivity of the output sentiment to protected attributes. Overall, we demonstrated that our rating system can help users compare alternatives to make informed decision regarding selection of SASs with respect to gender and racial-bias. Our method could also detect different in behavior when the output of SAS is discretized v/s left unchanged as continuous.


Our work can be extended by not only  relaxing the assumptions made but also trying them on more general settings.  
First, one could extend it to more complex input - i.e., sentence templates with more that one emotion word and gender / race variables in each sentence. 
Second, the proposed rating methodology, with little tweaks, can be applied to other AI systems like machine translators and object detectors which too  have exhibited the bias issue. Third, one could explore composite settings 
where diverse types of AIs are used together like SAS and machine translator, e.g., (\cite{round-trip}.


\bibliography{references}

%
%








\onecolumn

\aistatstitle{Supplementary Material for \\ `Rating Sentiment Analysis Systems for Bias through a Causal Lens'}

In this supplementary material, we provide additional information about related work,  experiments and results to help better understand our work in detail. The material is organized as follows:




\appendix

\section{Related Work}

\subsection{Adversarial Attacks Using Sentiments}

The instability of SAS is often a source of adversarial attacks in AI systems. This can manifest as adversarial examples (\cite{adv-example}), which modify test
inputs into a model to cause it to produce incorrect outputs (sentiments for SAS), or 
backdoor attacks (\cite{backdoor}) which compromise the model by poisoning the training data  and/or modifying the training. A more sophisticated form of attack is proposed in
(\cite{spin-attack}) where a high-dimensional AI  model, like for  summarization, outputs positive summaries of any text that
mentions the name of some “meta-backdoor" like individual or organization.

\subsection{Handling Bias}

There have been prior efforts to address bias by improving handling of training data, improved methods or post-processing on the AI system's output (\cite{bias-survey}).
However, the research on bias in NLP systems has shortcomings. In (\cite{prob-bias-text}), the authors 
report that quantitative techniques for measuring or mitigating ``bias" are poorly matched to their motivations and argue for the need to conduct studies in the social context of the real world. Furthermore, work is needed to align technical definitions with what is legally enforceable (\cite{bias-definitions-legal}). 
In this regard, to convey the behavior of AI to stakeholders in a better way, a new idea was introduced to test and rate AI services which we discuss next.

\section{Experiments and Results}

\subsection{Group-2:}

{\bf Hypothesis:} Would gender affect the sentiment values computed by the SASs when there is a possibility of confounding effect?


{\bf Experimental Setup:} Here, we do not calculate the sentiment distribution across each of the protected attribute classes as the emotion word distribution for each of the genders is different. Instead, we use our proposed DIE \% to compare the sentiment distribution before and after deconfounding as the presence of confounder opens a backdoor path from the input to the output through the confounders. The causal link between \emph{Gender} and \emph{Emotion Word} denotes that the gender affects the way emotion words are associated with a specific gender. For example, in one of the cases, we associate positive words more with sentences having a specific gender variable than sentences with other gender variables. We select one of these cases to illustrate the results. For example, in some scenarios, positive emotion words might be associated more with sentences with male gender variable. The choice of the case depends on the socio-technical context to be studied; our experimental setting is motivated by the results from literature for face detection algorithms where error rates for African-American females were found to be disproportionately higher than that for European males (\cite{gender-shades-image}). The authors want to emphasize that they do not endorse any form of bias in society or systemic errors in AI algorithms. In fact, that is the reason to test and rate AI systems.  Our method would work for any gender combination to be evaluated.  Table \ref{tab:emo-dist} shows different cases that can be considered. k=0 represents the scenario where the emotion words are associated uniformly across all the genders. We choose k=1 for this experiment in which 90\% of the sentences with male gender variables are associated with positive emotion words and rest with negative emotion words. For sentences with female gender variables, it is vice-versa and for NA, emotion words are equally distributed for male and female. This makes the number of positive and negative emotion words in the distribution equal. Table \ref{tab:g2} shows the results of this experiment when the sentiment is discretized and Table \ref{tab:g2-cont} shows the results when the original sentiment values are used without discretizing.

\begin{table}[h]
{\small
\centering
\begin{tabular}{ | m{1cm} | m{1cm}| m{1cm} | m{1em} | } 
  \hline
  {\bf Male} & {\bf Female} & {\bf NA} & {\bf k}  \\ 
  \hline
  \hline
  (50,50) & (50,50) & (50,50)  & 0 \\ 
  \hline
  (90,10) & (10,90) & (50,50)  & 1 \\ 
  \hline
  (10,90) & (90,10) & (50,50)  & 2 \\ 
  \hline
  (90,10) & (50,50) & (10,90)  & 3 \\ 
  \hline
  (10,90) & (50,50) & (90,10)  & 4 \\ 
  \hline
  (50,50) & (90,10) & (10,90)  & 5 \\
  \hline
  (50,50) & (10,90) & (90,10)  & 6 \\ 
  \hline
\end{tabular}
\caption{Different dataset distributions that can be considered holding the number of positive and negative words constant in each of the cases along with the number of male, female and NA sentences. Each of these tuples represent the \% of (positive,negative) emotion words associated with that particular gender. 'k' is the label that is used to represent each of these cases.}
\label{tab:emo-dist}
}
\end{table}

\begin{table}[h]
\centering
   {\tiny
    \begin{tabular}{|p{3em}|p{3.5em}|p{5em}|p{6em}|p{5em}|p{4em}|}
    \hline
    
          {\bf SAS} &    
          {\bf E.words} &
          {\bf E[\emph{Sentiment $|$ Emotion Word}]} &
          {\bf E[\emph{Sentiment $|$ do(Emotion Word)}]} &
          {\bf DIE \%} &
          {\bf MAX(DIE \%)}
          \\ \hline 
          
          $S_b$ & 
          E3 &
          (0.23,-1)   &
          (-0.08,-0.24)   &
          (134.7, 76)&
          76
          \\ \hline
          & 
          E4 &
          (0.40,-0.85)   &
          (0.02,-0.16)   &
          (95,81.17)   &
          95
          \\ \hline
          & 
          E5 &
          (0.14,-1)   &
          (-0.04,-0.28)   &
          (128.5, 72)   &
          128.5
          \\ \hline
          
          $S_r^\dagger$ & 
          E3 &
          (0.54,0.38)   &
          (0.55,0.40)   &
          (1.85,5.26) &
          5.26
          \\ \hline
          & 
          E4 &
          (0.46,0.72)   &
          (0.41,0.77)   &
          (10.87,6.94) &
          10.87$^*$
          \\ \hline
          & 
          E5 &
          (0.57,0.4)   &
          (0.57,0.4)   &
          (0,0) &
          0
          \\ \hline
          
          $S_t^\dagger$ & 
          E3 &
          (0,1)   &
          (0,1)   &
          (0,0) &
          0
          \\ \hline
          & 
          E4 &
          (0,1)   &
          (0,1)   &
          (0,0) &
          0
          \\ \hline
          & 
          E5 &
          (0,1)   &
          (0,1)   &
          (0,0) &
          0
          \\ \hline
          
          $S_d^\dagger$ & 
          E3 &
          (0,1)   &
          (0,1)   &
          (0,0) &
          0
          \\ \hline
          & 
          E4 &
          (0,1)   &
          (0,1)   &
          (0,0) &
          0
          \\ \hline
          & 
          E5 &
          (0,1)   &
          (0,1)   &
          (0,0) &
          0
          \\ \hline
          
          $S_g^\dagger$ & 
          E3 &
          (0.08,0.33)   &
          (0.09,0.30)   &
          (12.50,9.09) &
          12.50
          \\ \hline
          & 
          E4 &
          (0.06,0.47)   &
          (0.07,0.45)   &
          (16.66,4.25) &
          16.66$^*$
          \\ \hline
          & 
          E5 &
          (0.01,0.49)   &
          (0,0.5)   &
          (X,2.04) &
          X$^*$
          \\ \hline
        
    \end{tabular}
        \caption{E[Sentiment $|$ Emotion Word] and E[Sentiment $|$ do(Emotion Word)] values for Group 2 datasets when output sentiment is discretized and the DIE \% when emotion word sets, E3, E4 and E5 are considered. We then compute the MAX() from the DIE \%.}
    \label{tab:g2}
    }
\end{table}

\begin{table}
\centering
{\tiny
    \begin{tabular}{|p{2em}|p{3em}|p{7em}|p{7em}|p{6em}|p{4em}|}
    \hline
    
          {\bf SAS} &    
          {\bf E.words} &
          {\bf E[\emph{Sentiment $|$ Emotion Word}]} &
          {\bf E[\emph{Sentiment $|$ do(Emotion Word)}]} &
          {\bf DIE \%} & 
          {\bf MAX(DIE \%)}
          \\ \hline 
          
          $S_b$ & 
          E3 &
          (0.23,-1)   &
          (-0.08,-0.24)   &
          (134.7, 76)&
          76
          \\ \hline
          & 
          E4 &
          (0.40,-0.85)   &
          (0.02,-0.16)   &
          (95,81.17)   &
          95
          \\ \hline
          & 
          E5 &
          (0.14,-1)   &
          (-0.04,-0.28)   &
          (128.5, 72)   &
          128.5
          \\ \hline
          
          $S_r$ & 
          E3 &
          (0.12,0.28)   &
          (0.17,0.28)   &
          (41.66,0) &
          41.66
          \\ \hline
          & 
          E4 &
          (-0.20,0.09)   &
          (-0.09,0.06)   &
          (55, 33.33)     &
          55
          \\ \hline
          & 
          E5 &
          (-0.27,0.07)   &
          (-0.22,0.02)   &
          (18.51,71.43)   &
          71.43
          \\ \hline
          
          $S_t$ & 
          E3 &
          (-1,0.80)   &
          (-0.24,0.58)   &
          (76, 27.50) &
          76
          \\ \hline
          & 
          E4 &
          (-0.81,0.80)   &
          (-0.81,0.80)   &
          (0,0)   &
          0
          \\ \hline
          & 
          E5 &
          (-1,0.65)   &
          (-1,0.65)   &
          (0,0)   &
          0
          \\ \hline
          
          $S_d$ & 
          E3 &
          (-1,1)   &
          (-0.26,0.78)   &
          (74, 22)   &
          74
          \\ \hline
          & 
          E4 &
          (-1,1)   &
          (-0.16,0.88)   &
          (84,12)   &
          84
          \\ \hline
          & 
          E5 &
          (-1,1)   &
          (-0.26,0.76)   &
          (74,24)   &
          74
          \\ \hline
        
         $S_g$ & 
          E3 &
          (-0.42,-0.15)   &
          (-0.43,-0.13)   &
          (2.38, 13.33)   &
          13.33
          \\ \hline
          & 
          E4 &
          (-0.45,-0.07)   &
          (-0.46,-0.06)   &
          (2.17,14.28)   &
          14.28
          \\ \hline
          & 
          E5 &
          (-0.40,-0.07)   &
          (-0.39,-0.04)   &
          (2.50, 42.85)   &
          42.85
          \\ \hline
        
    \end{tabular}
    \caption{E[Sentiment $|$ Emotion Word] and E[Sentiment $|$ do(Emotion Word)] values for Group 2 datasets when the original sentiment values (not discretized) are used and the DIE \% when emotion word sets, E3, E4 and E5 are considered. We then compute the MAX() from the DIE \%.}
    \label{tab:g2-cont}
    }
\end{table}

\subsection{Group-3:}
{\bf Hypothesis:} Would \emph{Gender} and \emph{Race} affect the sentiment value computed by the SASs when there is no possibility of confounding effect?

{\bf Experimental Setup:} The setup is similar to what we used for Group-1. However, there is one extra protected attribute in this case. We compare gender distributions and race distributions separately. Table \ref{tab:g3} shows the results of this experiment when the sentiment is discretized and Table \ref{tab:g3-cont} shows the results when sentiment is not discretized. They show the t-values obtained from each of the SASs for each emotion word set and the superscripts for the different values indicate the CIs for which the null hypothesis is rejected or accepted as described in Group-1 experiments.

$R_e$$R_n$: Absolute t-value computed for European and NA distributions.

$R_e$$R_a$: Absolute t-value computed for European and African-American distributions.

$R_a$$R_n$: Absolute t-value computed for African-American and NA distributions.

\begin{table}
\centering
   {\tiny
    \begin{tabular}{|p{4em}|p{2em}|p{3em}|p{3em}|p{3em}|p{3em}|p{3em}|p{3em}|}
    \hline
          {\bf SAS} &    
          {\bf E. words} & 
          {\bf $G_m$$G_n$} &
          {\bf $G_m$$G_f$} &
          {\bf $G_f$$G_n$} &
          {\bf $R_e$$R_n$} &
          {\bf $R_e$$R_a$} &
          {\bf $R_a$$R_n$}
          \\ \hline 
                  $S_b$ & 
          E1 &
          0 &
          H$^1$ &
          H$^1$ &
          2.64$^1$ &
          0 &
          2.64$^1$
         \\ \hline 
         & 
          E2 &
          0 &
          H$^1$ &
          H$^1$&
          2.64$^1$ &
          0 &
          2.64$^1$
          \\\hline
          & 
          E3 &
          0 &
          H$^1$ &
          H$^1$ &
          3.87$^1$ &
          0 &
          3.87$^1$
          \\ \hline
          & 
          E4 &
          0 &
          H$^1$ &
          H$^1$ &
          4.80$^1$ &
          0 &
          4.80$^1$
          \\ \hline
          & 
          E5 &
          0 &
          H$^1$ &
          H$^1$ &
          4.80$^1$ &
          0 &
          4.80$^1$
          \\ \hline
          
          $S_r^\dagger$ & 
          E1 &
          0 &
          0.97 &
          0.97 &
          0.48     &
          0 &
          0.48
          \\\hline 
         & 
         E2 &
          1.53$^2$ &
          0.50 &
          0.97 &
          0.48 &
          1.65$^2$ &
          2.25$^2$
          \\
        
         \hline
          
          & 
          E3 &
          0.34 &
          0.34 &
          0 &
          0.69 &
          1.04 &
          0.35
          \\ \hline

          & 
          E4 &
          0.57 &
          0.28 &
          0.85 &
          1.77$^2$ &
          2.08$^2$   &
          0.28
          \\ \hline
          
          & 
          E5 &
          0.29 &
          1.44$^3$ &
          1.15 &
          0.28 &
          0.28 &
          0.56
          \\ \hline

          $S_t^\dagger$ & 
          E1 &
          0 &
          0 &
          0 &
          0 &
          0 &
          0
          \\

         \hline 
          
         & 
          E2 &
          0 &
          0 &
          0 &
          0 &
          0 &
          0
          \\
        
         \hline
          
          & 
          E3 &
          0 &
          0 &
          0 &
          0 &
          0 &
          0
          \\ \hline

          & 
          E4 &
          0 &
          0 &
          0 &
          0 &
          0 &
          0
          \\ \hline
          
          & 
          E5 &
          0 &
          0 &
          0 &
          0 &
          0 &
          0
                    \\ \hline
          $S_d^\dagger$ & 
          E1 &
          0 &
          0 &
          0 &
          0 &
          0 &
          0
          \\

         \hline 
          
         & 
          E2 &
          0 &
          0 &
          0 &
          0 &
          0 &
          0
          \\
        
         \hline
          
          & 
          E3 &
          0 &
          0 &
          0 &
          0 &
          0 &
          0
          \\ \hline

          & 
          E4 &
          0 &
          0 &
          0 &
          0 &
          0 &
          0
          \\ \hline
          
          & 
          E5 &
          0 &
          0 &
          0 &
          0 &
          0 &
          0
          \\ \hline
          
          $S_g^\dagger$ & 
          E1 &
          0 &
          0 &
          0 &
          0 &
          0 &
          0
          \\
         \hline 
          
         & 
          E2 &
          1 &
          0 &
          1 &
          1 &
          0 &
          1
          \\
        
         \hline
          
          & 
          E3 &
          0.89 &
          0 &
          0.89 &
          0.89 &
          0 &
          0.89
          \\ \hline
          & 
          E4 &
          1.55$^2$ &
          0 &
          1.55$^2$ &
          1.55$^2$ &
          0 &
          1.55$^2$
          \\ \hline
          
          & 
          E5 &
          1.33$^3$ &
          0 &
          1.33$^3$ &
          1.33$^3$ &
          0 &
          1.33$^3$
          \\ \hline
    \end{tabular}
    \caption{Results for Group 3 datasets (when output sentiment is discretized) showing the t-values and the superscript shows whether the null hypothesis is rejected or accepted in each case for the CIs considered (95\%, 70\%, 60\%). Superscript '1' indicates rejection with all 3 CIs, '2' indicates rejection with 70 and 60. '3' indicates rejection with 60 \%.}
    \label{tab:g3}
    }
\end{table}

\begin{table}
\centering
   {\tiny
    \begin{tabular}{|p{4em}|p{2em}|p{3em}|p{3em}|p{3em}|p{3em}|p{3em}|p{3em}|}
    \hline
          {\bf SAS} &    
          {\bf E. words} & 
          {\bf $G_m$$G_n$} &
          {\bf $G_m$$G_f$} &
          {\bf $G_f$$G_n$} &
          {\bf $R_e$$R_n$} &
          {\bf $R_e$$R_a$} &
          {\bf $R_a$$R_n$}
          \\ \hline 
          $S_b$ & 
          E1 &
          0 &
          H$^1$ &
          H$^1$ &
          2.64$^1$ &
          0 &
          2.64$^1$
         \\ \hline 
         & 
          E2 &
          0 &
          H$^1$ &
          H$^1$&
          2.64$^1$ &
          0 &
          2.64$^1$
          \\\hline
          & 
          E3 &
          0 &
          H$^1$ &
          H$^1$ &
          3.87$^1$ &
          0 &
          3.87$^1$
          \\ \hline
          & 
          E4 &
          0 &
          H$^1$ &
          H$^1$ &
          4.80$^1$ &
          0 &
          4.80$^1$
          \\ \hline
          & 
          E5 &
          0 &
          H$^1$ &
          H$^1$ &
          4.80$^1$ &
          0 &
          4.80$^1$
          \\ \hline
          
          $S_r$ & 
          E1 &
          0.45 &
          0.79 &
          0.21 &
          0.03     &
          0.24 &
          0.23
          \\\hline 
         & 
         E2 &
          0.28 &
          0.97 &
          0.67 &
          0.42 &
          0.37 &
          0.03
          \\
        
         \hline
          
          & 
          E3 &
          0.80 &
          0.73 &
          1.60$^2$ &
          1.80$^2$ &
          1.12 &
          0.62
          \\ \hline

          & 
          E4 &
          1.66$^2$ &
          0.37 &
          1.91$^2$ &
          2.37$^2$ &
          0.99   &
          1.25
          \\ \hline
          
          & 
          E5 &
          2.15$^2$ &
          0.60 &
          2.97$^1$ &
          2.84$^1$ &
          0.47 &
          2.26$^2$
          \\ \hline

          $S_t$ & 
          E1 &
          0 &
          0 &
          0 &
          0 &
          0 &
          0
          \\

         \hline 
          
         & 
          E2 &
          0 &
          0 &
          0 &
          0 &
          0 &
          0
          \\
        
         \hline
          
          & 
          E3 &
          0 &
          0 &
          0 &
          0 &
          0 &
          0
          \\ \hline

          & 
          E4 &
          0 &
          0 &
          0 &
          0 &
          0 &
          0
          \\ \hline
          
          & 
          E5 &
          0 &
          0 &
          0 &
          0 &
          0 &
          0
                    \\ \hline
          $S_d$ & 
          E1 &
          0 &
          0 &
          0 &
          0 &
          0 &
          0
          \\

         \hline 
          
         & 
          E2 &
          0 &
          0 &
          0 &
          0 &
          0 &
          0
          \\
        
         \hline
          
          & 
          E3 &
          0 &
          0 &
          0 &
          0 &
          0 &
          0
          \\ \hline

          & 
          E4 &
          0 &
          0 &
          0 &
          0 &
          0 &
          0
          \\ \hline
          
          & 
          E5 &
          0 &
          0 &
          0 &
          0 &
          0 &
          0
          \\ \hline
          
          $S_g$ & 
          E1 &
          1 &
          0 &
          1 &
          1 &
          0 &
          1
          \\
         \hline 
          
         & 
          E2 &
          1.34 &
          0 &
          1.34 &
          1.34 &
          0 &
          1.34
          \\
        
         \hline
          
          & 
          E3 &
          0.69 &
          0 &
          0.69 &
          0.69 &
          0 &
          0.69
          \\ \hline
          & 
          E4 &
          0.57 &
          0 &
          0.57 &
          0.57 &
          0 &
          0.57
          \\ \hline
          
          & 
          E5 &
          0.94 &
          0 &
          0.94 &
          0.94 &
          0 &
          0.94
          \\ \hline
    \end{tabular}
    \caption{Results for Group 3 datasets (when the original sentiment values are used (not discretized)) showing the t-values and the superscript shows whether the null hypothesis is rejected or accepted in each case for the CIs considered (95\%, 70\%, 60\%). Superscript '1' indicates rejection with all 3 CIs, '2' indicates rejection with 70 and 60. '3' indicates rejection with 60 \%.}
    \label{tab:g3-cont}
    }
\end{table}

We also consider a composite case where we combine both race and gender attributes together to form one single attribute called \emph{RG} as shown in Table \ref{tab:g3-comp} and Table \ref{tab:g3-comp-cont} . For example, in the composite case, European name and male gender would be considered as a European male and we compute the pairwise t-values for each of these distributions. In the tables, the subscripts, `n' denotes NA, `em' denotes European male, `ef' denotes European female, `am' denotes African-American male, `af' denotes African-American female. 


\begin{table*}[!hbt]
\centering
  {\tiny
    \begin{tabular}{|p{2em}|p{2em}|p{5em}|p{5em}|p{5em}|p{5em}|p{6em}|p{6em}|p{6em}|p{6em}|p{6em}|p{6em}|}
    \hline
          {\bf SAS} &    
          {\bf E. words} & 
          {\bf $RG_n$$RG_{em}$} &
          {\bf $RG_n$$RG_{ef}$} &
          {\bf $RG_n$$RG_{am}$} &
          {\bf $RG_n$$RG_{af}$} &
          {\bf $RG_{em}$$RG_{ef}$} &
          {\bf $RG_{em}$$RG_{am}$} &
          {\bf $RG_{em}$$RG_{af}$} &
          {\bf $RG_{ef}$$RG_{am}$} &
          {\bf $RG_{ef}$$RG_{af}$} &
          {\bf $RG_{am}$$RG_{af}$} 
          \\ \hline 
         $S_b$ & 
          E1 &
          0 &
          H$^1$ &
          0 &
          H$^1$ &
          H$^1$ &
          0 &
          H$^1$ &
          H$^1$ &
          0 &
          H$^1$
          \\\hline 
         & 
          E2 &
          0 &
          H$^1$ &
          0 &
          H$^1$ &
          H$^1$ &
          0 &
          H$^1$ &
          H$^1$ &
          0 &
          H$^1$
          \\\hline
          & 
          E3 &
          0 &
          H$^1$ &
          0 &
          H$^1$ &
          H$^1$ &
          0 &
          H$^1$ &
          H$^1$ &
          0 &
          H$^1$
          \\ \hline
          & 
          E4 &
          0 &
          H$^1$ &
          0 &
          H$^1$ &
          H$^1$ &
          0 &
          H$^1$ &
          H$^1$ &
          0 &
          H$^1$
          \\ \hline
          & 
          E5 &
          0 &
          H$^1$&
          0 &
          H$^1$ &
          H$^1$&
          0 &
          H$^1$ &
          H$^1$ &
          0 &
          H$^1$
          \\ \hline
          
          $S_r^\dagger$ & 
          E1 &
          0.40 &
          1.21 &
          0.37 &
          0.37 &
          1.41 &
          0.65 &
          0.65 &
          0.65 &
          0.65 &
          0
          \\ \hline 
          
         & 
         E2 &
          1.21 &
          0.40 &
          1.21 &
          3.41$^1$ &
          1.41 &
          0 &
          1 &
          1.41 &
          3$^1$ &
          1
          \\\hline
          
          & 
          E3 &
          0.84 &
          0.28 &
          0.28 &
          0.28 &
          0.48 &
          0.97 &
          0.97 &
          0.48 &
          0.48 &
          0
          \\ \hline

          & 
          E4 &
          1.18 &
          1.75$^2$ &
          0.23 &
          0.23 &
          0.43 &
          1.21 &
          1.21 &
          1.68$^2$ &
          1.68$^2$ &
          0
          \\ \hline
          
          & 
          E5 &
          0 &
          0.46 &
          0.47 &
          1.42$^3$ &
          0.40 &
          0.40 &
          1.21&
          0.80 &
          0.80 &
          1.66$^2$  
          \\ \hline
          
          $S_t^\dagger$ & 
          E1 &
          0 &
          0 &
          0 &
          0 &
          0 &
          0 &
          0 &
          0 &
          0 &
          0
          \\\hline 
         & 
          E2 &
          0 &
          0 &
          0 &
          0 &
          0 &
          0 &
          0 &
          0 &
          0 &
          0
          \\\hline
          & 
          E3 &
          0 &
          0 &
          0 &
          0 &
          0 &
          0 &
          0 &
          0 &
          0 &
          0
          \\ \hline
          & 
          E4 &
          0 &
          0 &
          0 &
          0 &
          0 &
          0 &
          0 &
          0 &
          0 &
          0
          \\ \hline
          & 
          E5 &
          0 &
          0 &
          0 &
          0 &
          0 &
          0 &
          0 &
          0 &
          0 &
          0
        \\ \hline
          $S_d^\dagger$ & 
          E1 &
          0 &
          0 &
          0 &
          0 &
          0 &
          0 &
          0 &
          0 &
          0 &
          0
          \\\hline 
         & 
          E2 &
          0 &
          0 &
          0 &
          0 &
          0 &
          0 &
          0 &
          0 &
          0 &
          0
          \\\hline
          & 
          E3 &
          0 &
          0 &
          0 &
          0 &
          0 &
          0 &
          0 &
          0 &
          0 &
          0
          \\ \hline
          & 
          E4 &
          0 &
          0 &
          0 &
          0 &
          0 &
          0 &
          0 &
          0 &
          0 &
          0
          \\ \hline
          & 
          E5 &
          0 &
          0 &
          0 &
          0 &
          0 &
          0 &
          0 &
          0 &
          0 &
          0
          \\ \hline
          
          $S_g^\dagger$ & 
          E1 &
          0 &
          0 &
          0 &
          0 &
          0 &
          0 &
          0 &
          0 &
          0 &
          0
          \\\hline 
         & 
          E2 &
          0.75 &
          0.75 &
          0.75 &
          0.75 &
          0 &
          0 &
          0 &
          0 &
          0 &
          0
          \\\hline
          & 
          E3 &
          0.68 &
          0.68 &
          0.68 &
          0.68 &
          0 &
          0 &
          0 &
          0 &
          0 &
          0
          \\ \hline
          & 
          E4 &
          1.17 &
          1.17 &
          1.17 &
          1.17 &
          0 &
          0 &
          0 &
          0 &
          0 &
          0
           \\ \hline
          & 
          E5 &
          1.03 &
          1.03 &
          1.03 &
          1.03 &
          0 &
          0 &
          0 &
          0 &
          0 &
          0
          \\ \hline

    \end{tabular}
    \caption{Results for Group 3 composite case datasets (when output sentiment values are discretized) showing the t-values and the superscript shows whether the null hypothesis is rejected or accepted in each case for the CIs considered (95\%, 70\%, 60\%). Superscript '1' indicates rejection with all 3 CIs, '2' indicates rejection with 70 and 60. '3' indicates rejection with 60 \%.}    
    \label{tab:g3-comp}
    }
\end{table*}

\begin{table*}[!hbt]
\centering
  {\tiny
    \begin{tabular}{|p{2em}|p{2em}|p{5em}|p{5em}|p{5em}|p{5em}|p{6em}|p{6em}|p{6em}|p{6em}|p{6em}|p{6em}|}
    \hline
          {\bf SAS} &    
          {\bf E. words} & 
          {\bf $RG_n$$RG_{em}$} &
          {\bf $RG_n$$RG_{ef}$} &
          {\bf $RG_n$$RG_{am}$} &
          {\bf $RG_n$$RG_{af}$} &
          {\bf $RG_{em}$$RG_{ef}$} &
          {\bf $RG_{em}$$RG_{am}$} &
          {\bf $RG_{em}$$RG_{af}$} &
          {\bf $RG_{ef}$$RG_{am}$} &
          {\bf $RG_{ef}$$RG_{af}$} &
          {\bf $RG_{am}$$RG_{af}$} 
          \\ \hline 
          $S_b$ & 
          E1 &
          0 &
          H$^1$ &
          0 &
          H$^1$ &
          H$^1$ &
          0 &
          H$^1$ &
          H$^1$ &
          0 &
          H$^1$
          \\\hline 
         & 
          E2 &
          0 &
          H$^1$ &
          0 &
          H$^1$ &
          H$^1$ &
          0 &
          H$^1$ &
          H$^1$ &
          0 &
          H$^1$
          \\\hline
          & 
          E3 &
          0 &
          H$^1$ &
          0 &
          H$^1$ &
          H$^1$ &
          0 &
          H$^1$ &
          H$^1$ &
          0 &
          H$^1$
          \\ \hline
          & 
          E4 &
          0 &
          H$^1$ &
          0 &
          H$^1$ &
          H$^1$ &
          0 &
          H$^1$ &
          H$^1$ &
          0 &
          H$^1$
          \\ \hline
          & 
          E5 &
          0 &
          H$^1$&
          0 &
          H$^1$ &
          H$^1$&
          0 &
          H$^1$ &
          H$^1$ &
          0 &
          H$^1$
          \\ \hline
          
          $S_r$ & 
          E1 &
          0.20 &
          0.24 &
          0.53 &
          0.14 &
          0.43 &
          0.27 &
          0.30 &
          0.83 &
          0.03 &
          0.60
          \\ \hline 
          
         & 
         E2 &
          0.06 &
          0.66 &
          0.37 &
          0.37 &
          0.68 &
          0.30 &
          0.40 &
          0.87 &
          0.20 &
          0.61 
          \\\hline
          
          & 
          E3 &
          1.73$^2$ &
          1.13 &
          0.54 &
          1.41$^3$ &
          0.63 &
          2.04$^2$ &
          0.25 &
          1.51$^3$ &
          0.35 &
          1.75$^2$
          \\ \hline

          & 
          E4 &
          2.54$^1$ &
          1.41$^3$ &
          0.40 &
          1.57$^2$ &
          0.53 &
          1.66$^2$ &
          0.46 &
          0.91 &
          0.07 &
          1.02
          \\ \hline
          
          & 
          E5 &
          1.46$^3$ &
          2.94$^1$ &
          1.73$^2$ &
          1.61$^2$ &
          0.80 &
          0.05 &
          0.07 &
          0.81 &
          0.73 &
          0.03   
          \\ \hline
          
          $S_t$ & 
          E1 &
          0 &
          0 &
          0 &
          0 &
          0 &
          0 &
          0 &
          0 &
          0 &
          0
          \\\hline 
         & 
          E2 &
          0 &
          0 &
          0 &
          0 &
          0 &
          0 &
          0 &
          0 &
          0 &
          0
          \\\hline
          & 
          E3 &
          0 &
          0 &
          0 &
          0 &
          0 &
          0 &
          0 &
          0 &
          0 &
          0
          \\ \hline
          & 
          E4 &
          0 &
          0 &
          0 &
          0 &
          0 &
          0 &
          0 &
          0 &
          0 &
          0
          \\ \hline
          & 
          E5 &
          0 &
          0 &
          0 &
          0 &
          0 &
          0 &
          0 &
          0 &
          0 &
          0
        \\ \hline
          $S_d$ & 
          E1 &
          0 &
          0 &
          0 &
          0 &
          0 &
          0 &
          0 &
          0 &
          0 &
          0
          \\\hline 
         & 
          E2 &
          0 &
          0 &
          0 &
          0 &
          0 &
          0 &
          0 &
          0 &
          0 &
          0
          \\\hline
          & 
          E3 &
          0 &
          0 &
          0 &
          0 &
          0 &
          0 &
          0 &
          0 &
          0 &
          0
          \\ \hline
          & 
          E4 &
          0 &
          0 &
          0 &
          0 &
          0 &
          0 &
          0 &
          0 &
          0 &
          0
          \\ \hline
          & 
          E5 &
          0 &
          0 &
          0 &
          0 &
          0 &
          0 &
          0 &
          0 &
          0 &
          0
          \\ \hline
          
          $S_g$ & 
          E1 &
          0.75 &
          0.75 &
          0.75 &
          0.75 &
          0 &
          0 &
          0 &
          0 &
          0 &
          0
          \\\hline 
         & 
          E2 &
          1.06 &
          1.06 &
          1.06 &
          1.06 &
          0 &
          0 &
          0 &
          0 &
          0 &
          0
          \\\hline
          & 
          E3 &
          0.53 &
          0.53 &
          0.53 &
          0.53 &
          0 &
          0 &
          0 &
          0 &
          0 &
          0
          \\ \hline
          & 
          E4 &
          0.43 &
          0.43 &
          0.43 &
          0.43 &
          0 &
          0 &
          0 &
          0 &
          0 &
          0
           \\ \hline
          & 
          E5 &
          0.72 &
          0.72 &
          0.72 &
          0.72 &
          0 &
          0 &
          0 &
          0 &
          0 &
          0
          \\ \hline

    \end{tabular}
    \caption{Results for Group 3 composite case (when original sentiment values are used (not discretized)) datasets showing the t-values and the superscript shows whether the null hypothesis is rejected or accepted in each case for the CIs considered (95\%, 70\%, 60\%). Superscript '1' indicates rejection with all 3 CIs, '2' indicates rejection with 70 and 60. '3' indicates rejection with 60 \%.}    
    \label{tab:g3-comp-cont}
    }
\end{table*}

\begin{table}
\centering
   {\tiny
    \begin{tabular}{|p{4em}|p{2em}|p{3em}|p{3em}|p{3em}|}
    \hline
    
          \bf{SAS} &    
          \bf{E. words} & 
          $\bf{G_mG_n}$ &
          $\bf{G_mG_f}$ &
          $\bf{G_fG_n}$ 
          \\ \hline

          $S_b$ & 
          E1 &
          0 &
          H$^1$ &
          H$^1$ 
          \\

         \hline 
          
         & 
          E2 &
          0 &
          H$^1$ &
          H$^1$ 
          \\
        
         \hline
          
          & 
          E3 &
          0 &
          H$^1$ &
          H$^1$ 
          \\ \hline

          & 
          E4 &
          0 &
          H$^1$ &
          H$^1$
          \\ \hline
          
          & 
          E5 &
          0 &
          H$^1$ &
          H$^1$
          \\ \hline
           
          $S_r$ & 
          E1 &
          0.06 &
          0.89 &
          1 
          \\

         \hline 
          
         & 
         E2 &
          0.13 &
          0.39 &
          0.32 
          \\
        
         \hline
          
          & 
          E3 &
          0.65 &
          1.42 &
          0.85 
          \\ \hline

          & 
          E4 &
          0.32 &
          0.60 &
          0.94 
          \\ \hline
          
          & 
          E5 &
          2.08$^2$ &
          1.12 &
          1.20 
          \\ \hline

          $S_t$ & 
          E1 &
          0 &
          0 &
          0 
          \\

         \hline 
          
         & 
          E2 &
          0 &
          0 &
          0 
          \\
        
         \hline
          
          & 
          E3 &
          0 &
          0 &
          0 
          \\ \hline

          & 
          E4 &
          0 &
          0 &
          0 
          \\ \hline
          
          & 
          E5 &
          0 &
          0 &
          0 
                    \\ \hline
          $S_d$ & 
          E1 &
          0 &
          0 &
          0 
          \\

         \hline 
          
         & 
          E2 &
          0 &
          0 &
          0 
          \\
        
         \hline
          
          & 
          E3 &
          0 &
          0 &
          0 
          \\ \hline

          & 
          E4 &
          0 &
          0 &
          0
          \\ \hline
          
          & 
          E5 &
          0 &
          0 &
          0 
          \\ \hline

          $S_g$ & 
          E1 &
          0.17 &
          0.17 &
          0
          \\
         \hline 
          
         & 
          E2 &
          1.15    &
          0.40 &
          0.90 
          \\
        
         \hline
          
          & 
          E3 &
          0.63 &
          0.12 &
          0.77 
          \\ \hline
          
          & 
          E4 &
          0.61 &
          0.10 &
          0.81  
          \\ \hline
          
          & 
          E5 &
          1.31$^3$ &
          0.64     &
          0.92 
          \\ \hline

    \end{tabular}
    \caption{Results for Group 1 datasets (when the original sentiment values (not discretized) are used) showing the t-values and the superscript shows whether the null hypothesis is rejected or accepted in each case for the CIs considered (95\%, 70\%, 60\%). Superscript '1' indicates rejection with all 3 CIs, '2' indicates rejection with 70 and 60. '3' indicates rejection with 60 \%.}
    \label{tab:g1-cont}
    }
\end{table}

\begin{table}
\centering
    \begin{tabular}{|p{2em}|p{3em}|p{7em}|p{7em}|p{6em}|p{4em}|}
    \hline
    
          {\bf SAS} &    
          {\bf E.words} &
          {\bf E[\emph{Sentiment $|$ Emotion Word}]} &
          {\bf E[\emph{Sentiment $|$ do(Emotion Word)}]} &
          {\bf DIE \%} & 
          {\bf MAX(DIE \%)}
          \\ \hline 
          
          $S_b$ & 
          E3 &
          (-0.16,-0.50)   &
          (-0.08,-0.08)   &
          (50,84)&
          84
          \\ \hline
          & 
          E4 &
          (-0.20,-0.55)   &
          (-0.10,0.03)   &
          (50,105.4)   &
          105.4
          \\ \hline
          & 
          E5 &
          (0.11,-0.60)   &
          (0.03,-0.11)   &
          (72.72,74.24)   &
          74.24
          \\ \hline
          
          $S_r$ & 
          E3 &
          (0.11,0.20)   &
          (0.13,0.29)   &
          (18.18,45) &
          45
          \\ \hline
          & 
          E4 &
          (-0.28,0.27)   &
          (-0.37,0.35)   &
          (32.14,29.62)     &
          32.14
          \\ \hline
          & 
          E5 &
          (-0.39,-0.15)   &
          (-0.43,-0.16)   &
          (10.25,6.66)   &
          10.25
          \\ \hline
          
          $S_t$ & 
          E3 &
          (-1,0.80)   &
          (-0.22,0.58)   &
          (78, 27.5) &
          78
          \\ \hline
          & 
          E4 &
          (-0.81,0.80)   &
          (-0.81,0.80)   &
          (0,0)   &
          0
          \\ \hline
          & 
          E5 &
          (-1,0.59)   &
          (-1,0.59)   &
          (0,0)   &
          0
          \\ \hline

          $S_d$ & 
          E3 &
          (-1,1)   &
          (-0.24,0.77)   &
          (76,23)   &
          76
          \\ \hline
          & 
          E4 &
          (-1,1)   &
          (-0.28,0.78)   &
          (72,22)   &
          72
          \\ \hline
          & 
          E5 &
          (-1,1)   &
          (-0.20,0.80)   &
          (80,20)   &
          80
          \\ \hline
          
          $S_g$ & 
          E3 &
          (-0.42,-0.07)   &
          (-0.42,-0.05)   &
          (0,28.57)  &
          28.57
          \\ \hline
          & 
          E4 &
          (-0.46,-0.07)   &
          (-0.45,-0.06)   &
          (0,14.28)   &
          14.28
          \\ \hline
          & 
          E5 &
          (-0.45,-0.16)   &
          (-0.45,-0.15)   &
          (0,6.25)   &
          6.25
          \\ \hline
        
    \end{tabular}
    \caption{E[Sentiment $|$ Emotion Word] and E[Sentiment $|$ do(Emotion Word)] values for Group 4 datasets when the original output sentiment values (not discretized) are used and the DIE \% when emotion word sets, E3, E4 and E5 are considered. We then compute the MAX() from the DIE \%.}
    \label{tab:g4-cont}
\end{table}



\end{document}